\documentclass[wcp]{jmlr}
\usepackage{microtype}
\usepackage{multirow, mathtools}
\usepackage{graphicx}
\usepackage{exscale,relsize}
\usepackage{inconsolata}
\usepackage{booktabs} 
\usepackage{diagbox}
\usepackage{todonotes}
\usepackage{hyperref}
\newcommand{\name}{AMP~}
\usepackage{amssymb}
\usepackage{pifont}
\newcommand{\cmark}{\ding{51}}%
\newcommand{\xmark}{\ding{55}}
\usepackage{listings}
\usepackage{derivative}

\usepackage{amsmath}
\usepackage{mathtools}
\usepackage{wrapfig,lipsum}

\usepackage[capitalize,noabbrev]{cleveref}
\crefname{section}{Sec.}{Secs.}
\Crefname{section}{Section}{Sections}
\Crefname{table}{Table}{Tables}
\crefname{table}{Tab.}{Tabs.}
\Crefname{figure}{Figure}{Figures}
\crefname{figure}{Fig.}{Figs.}
\usepackage{color}
\definecolor{deepblue}{rgb}{0,0,0.5}
\definecolor{deepred}{rgb}{0.6,0,0}
\definecolor{deepgreen}{rgb}{0,0.5,0}

\DeclareFixedFont{\ttb}{T1}{txtt}{bx}{n}{8} 
\DeclareFixedFont{\ttm}{T1}{txtt}{m}{n}{8}  
\newcommand\pythonstyle{\lstset{
language=Python,
basicstyle=\ttm,
morekeywords={self},              
keywordstyle=\ttb\color{deepblue},
emph={MyClass,__init__},          
emphstyle=\ttb\color{deepred},    
stringstyle=\color{deepgreen},
frame=tb,                         
showstringspaces=false
}}

\lstnewenvironment{python}[1][]
{
\pythonstyle
\lstset{#1}
}
{}

\makeatletter
\let\Ginclude@graphics\@org@Ginclude@graphics 
\makeatother

\jmlrvolume{189}
\jmlryear{2022}
\jmlrworkshop{ACML 2022}

\title[Out of Distribution Detection via Neural Network Anchoring]{Out of Distribution Detection via Neural Network Anchoring}
  
  \author{\Name{Rushil Anirudh} \Email{anirudh1@llnl.gov}\\
   \Name{Jayaraman J. Thiagarajan} \Email{jjayaram@llnl.gov}\\
    \addr Center for Applied Scientific Computing (CASC), \\ Lawrence Livermore National Laboratory}

\editors{Emtiyaz Khan and Mehmet G\"{o}nen}

\begin{document}

\maketitle

\begin{abstract}
 Our goal in this paper is to exploit heteroscedastic temperature scaling as a calibration strategy for out of distribution (OOD) detection. Heteroscedasticity here refers to the fact that the optimal temperature parameter for each sample can be different, as opposed to conventional approaches that use the same value for the entire distribution. To enable this, we propose a new training strategy called anchoring that can estimate appropriate temperature values for each sample, leading to state-of-the-art OOD detection performance across several benchmarks. Using NTK theory, we show that this temperature function estimate is closely linked to the epistemic uncertainty of the classifier, which explains its behavior. 
 In contrast to some of the best-performing OOD detection approaches, our method does not require exposure to additional outlier datasets, custom calibration objectives, or model ensembling. Through empirical studies with different OOD detection settings -- far OOD, near OOD, and semantically coherent OOD - we establish a highly effective OOD detection approach.  Code to reproduce our results is available at \href{https://github.com/LLNL/AMP}{github.com/LLNL/AMP}
\end{abstract}

\section{Introduction}

The task of using a trained model to accurately distinguish between samples from the dataset used for training -- \textit{i.e.}, the in-distribution (ID), and any other external dataset with different semantic characteristics is broadly referred to as OOD (out-of-distribution) detection. To solve this challenging problem, one needs to obtain an effective characterization of the ID data manifold, such that the discrepancy between test data and the \textit{inferred manifold} can be used to recognize the model's lack of knowledge about OOD data. This is commonly achieved by learning a scoring function: $\mathcal{S}: X \rightarrow \mathbb{R}$ that can score both ID and OOD samples appropriately. A simple scoring function can be based on the maximum softmax probability (MSP) of a prediction, with the expectation that the model will be more confident about an ID sample compared to OOD samples. However, in practice, such simple prediction confidence scores are poorly calibrated, and as a result, several novel scoring functions have emerged -- predictive entropy \citep{guo2017calibration}, energy~\citep{energyood}, uncertainty estimates~\citep{gal2016dropout, lakshminarayanan2016simple}, latent space deviation~\citep{van2020uncertainty}, class-specific deviations \citep{mahalanobis, sastry2020detecting} etc. Though these scoring functions often perform better than MSP, many state-of-the-art formulations~\citep{hendrycks18oe, energyood, scood} rely on additional unlabeled data for calibrating model predictions to better reject OOD data. In addition to requiring sophisticated training strategies (e.g., outlier exposure), this approach can be sub-optimal when the calibration dataset is not \textit{strictly} OOD, \textit{i.e.}, and they contain shared semantics with the ID set~\citep{scood}. Further, the calibration strategy used in many of these methods relies on \emph{temperature scaling} \citep{guo2017calibration}, which essentially scales the logits by a scalar called the temperature. When the temperature parameter is greater (or lower) than $1$, the entropy of the resulting prediction distribution increases (or decreases). Consequently, with an appropriate temperature value (chosen with either external or additional validation data), even this simple scaling leads to much improved OOD detection performance.


\paragraph{Heteroscedastic temperature scaling with anchoring.} In this paper, we explore the idea of \emph{heteroscedastic} temperature scaling, \textit{i.e.}, instead of using the same temperature scalar for all the samples, we construct a temperature function that produces sample-specific temperature values. Our hypothesis is that by appropriately tempering the predictions for ID and OOD samples, any existing scoring function can effectively between distinguish them. We achieve this using a novel training procedure called \emph{neural network anchoring}. In a nutshell, anchoring involves first transforming the input image, $\mathrm{x}$, into a tuple using the transformation $\mathcal{E}:\mathrm{x}\rightarrow [\mathrm{c}, \mathrm{x}-\mathrm{c}]$, where $\mathrm{c}$ is another randomly chosen image (``anchor'') from the training set, and predicting the label for $\mathrm{x}$ using this tuple. We also propose an additional consistency training strategy by perturbing the anchor before encoding, which boosts the performance further. During inference, we obtain predictions from multiple random anchors and propose to estimate the temperature based on standard deviation of these predictions. Using neural tangent kernel theory \citep{jacot2018neural}, we show that our heteroscedastic temperature estimate is closely related to the \emph{epistemic} uncertainty of the model.

\begin{figure*}[!htb]
    \centering
     \includegraphics[width=0.9\linewidth,trim=0cm 0cm 0cm 0cm,clip]{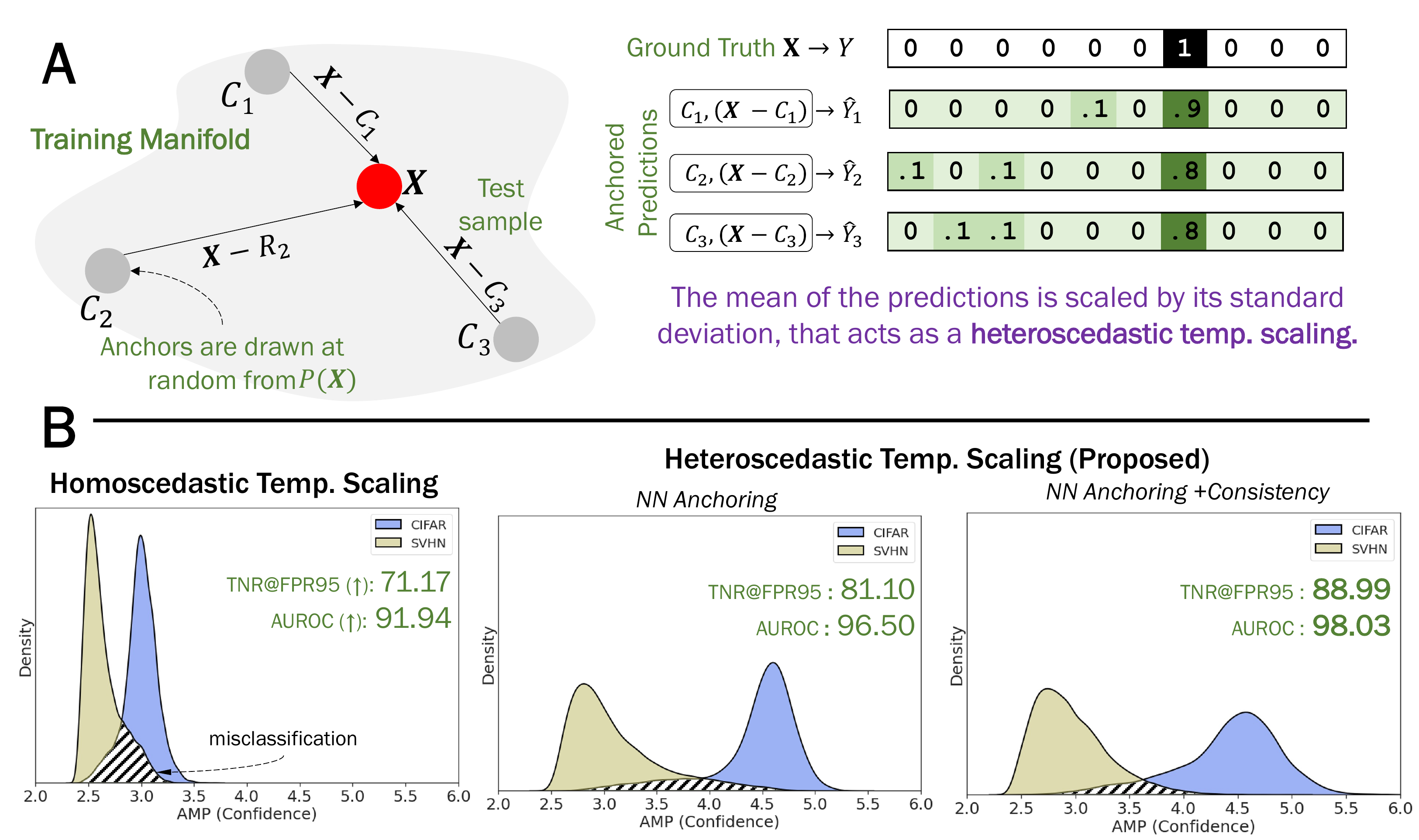}
     \vspace{-10pt}
    \caption{\small{\textbf{Improving existing OOD detectors via heteroscedastic temperature scaling.} (A) We propose a new training procedure called neural network anchoring to estimate the temperature parameter for any test sample, and show that it can be leveraged to improve conventional OOD detectors (e.g., entropy-based). (B) We also introduce a new consistency training objective to further improve the fidelity of detectors.}}
    \label{fig:anchoring_illus}
    \vspace{-10pt}
\end{figure*}

We use this temperature estimate to calibrate the predictions for a test sample, using which we can compute an OOD score using existing scoring functions (e.g., entropy). See \cref{fig:anchoring_illus}(A) for an illustration of the process, and \cref{fig:pseudocode,fig:pseudocode2} for the pseudo-codes. \cref{fig:anchoring_illus}(B) illustrates the improvement over conventional temperature scaling using the standard CIFAR-10/SVHN OOD benchmark.
Through extensive empirical analysis, we demonstrate that the proposed approach produces state-of-the-art detection performance across multiple benchmarks and models (summarized in Table \ref{table:summary}). 

\begin{table}[!htb]
\centering
\label{table:summary}
\footnotesize
\resizebox{\linewidth}{!}{
\begin{tabular}{c|l|c}
\toprule
 \textbf{Base Model}      & \textbf{Benchmark } (IN $\rightarrow$ OOD)\hfill Year & \textbf{Reference}           \\ \toprule
WRN-40-2 & CIFAR-10/100 $\rightarrow$ 6 Datasets \hfill(\citeauthor{odin}, ICLR'17)                & Table \ref{tab:ood_benchmark1}                                           \\ 
ResNet-34       & CIFAR-10/100 $\rightarrow$ 7 Datasets \hfill(\citeauthor{sastry2020detecting}, ICML'20) & Table \ref{tab:resnet34_ood}                                             \\ 
ResNet-34       & CIFAR-10 $\leftrightarrow$ CIFAR-100  \hfill(Near OOD)                                       & Table \ref{tab:near-ood}                                                  \\ 
ResNet-50       & ImageNet-1K $\rightarrow$ ImageNet-C \hfill(\citeauthor{krishnan2020improving}, NeurIPS'20)  & Table \ref{tab:Imagenet}, Figure \ref{fig:imagenet-ood} \\
ResNet-18       & Semantically Coherent OOD \hfill(\citeauthor{scood}, ICCV'21)                             & Table \ref{tab:scood}  \\ 
ResNet-34       & Robustness to resizing artifacts \hfill(this paper) & Table \ref{tab:resizing_ood} \\ 
 WRN-40-2& Ablation study & Table \ref{tab:ablations} \\\bottomrule
\end{tabular}
}
\caption{\small{Summary of experiments in this paper.}}
\vspace{-10pt}
\end{table}
\section{Background and Related Work}
We use training data $\mathcal{D} = \{(\mathrm{x}_i, y_i)\}_{i=1}^n$, where $\mathrm{x}_i \in P_I$ and $y_i \in \mathcal{C}_I \coloneqq \{1,2, \cdots, N_{class}\}$, to train a model $f(\boldsymbol{\theta}) \in \mathcal{H}$ with randomly initialized weights $\boldsymbol{\theta}_0$ and hypothesis space $\mathcal{H}$.%
We train a classifier $f(\theta): \mathrm{x}\rightarrow y$, parameterized by $\theta$ using $\mathcal{D}$. While the learned model is required to generalize to the test dataset $\mathcal{D}^{t} = \{(\mathrm{x}^t_i, y^t_i)\}_{i=1}^{N_t}$ when $\mathrm{x}^t_i \in P_I$ and $y^t_i \in \mathcal{C}_I$, it is also critical to recognize out-of-distribution samples, \textit{i.e.}, $\mathrm{x}^t_i \in P_O$ and $y^t_i \in \mathcal{C}_O$, on which the model could fail. Without loss of generality, this formulation encompasses scenarios with unknown distribution shifts to the input images, \textit{i.e.}, $P_O \neq P_I, \mathcal{C}_O \subseteq \mathcal{C}_I$ as well as the presence of additional, unknown classes $\mathcal{C}_O \supset \mathcal{C}_I$. 

\paragraph{Scoring functions} In summary, the goal of OOD detection is to use labeled data $\mathcal{D}$ to train a classifier that has the capacity to reject samples from $P_O$, while also accurately classifying samples from $P_I$. This is typically achieved by defining a scalar \emph{scoring function}, such that $\mathcal{S}_{\mathrm{x}\in P_I}\neq \mathcal{S}_{\mathrm{x}\in P_O}$, i.e., they are sufficiently distinct that an out of distribution sample is easily classified. Choosing the appropriate scoring function has been focus of the last several years of research, with most techniques choosing the score as some function of the prediction logits obtained from the trained classifier -- such as maximum softmax probability, entropy \citep{guo2017calibration}, and energy \citep{energyood}. 

\paragraph{Uncertainties for OOD detection} Uncertainty estimation has been a popular choice for OOD detection since epistemic (or model) uncertainties are supposed to be indicative of the OOD-ness of a test sample. For example, DUQ \citep{van2020uncertainty} uses a kernel distance to a set of class-specific centroids defined in the feature of a deep network as the measure for uncertainty. Another recent technique is DEUP \citep{jain2021deup} which trains an explicit epistemic uncertainty estimator for a pre-trained model. More generally, Bayesian methods \citep{neal2012bayesian} are among the most common kinds of uncertainty estimators today, but they are not easily scalable to large datasets and are known to be outperformed by model ensembling \citep{lakshminarayanan2016simple}. Monte Carlo Dropout \citep{gal2016dropout} is a scalable alternative to Bayesian methods in that it approximates the posterior distribution on the weights via dropout to estimate uncertainties.

\paragraph{Prediction calibration for OOD detection} In general uncertainty-based OOD detectors have so far not been able to improve performance over more traditional scoring based methods. As a result, a lot of focus has been on calibrating classifier predictions to make the scoring function more effective (often relying on outlier examples) such as Mahalanobis \citep{lee2018simple}, ODIN \citep{odin}, Gram matrices \citep{sastry2020detecting}, AVuC \citep{krishnan2020improving}, and outlier exposure \citep{hendrycks18oe}. The simplest, and often most effective, strategy for calibration is temperature scaling \citep{guo2017calibration} where the logits are scaled by a temperature parameter ($\tau$). When $\tau>1$ the prediction is made less confident (i.e., higher entropy), and if not it is made more peaky (i.e., lower entropy). Notably, most techniques define a single $\tau$ value to calibrate predictions on an entire test set. 

\paragraph{} The underlying assumption made by such approaches is that the predictions are \emph{homoscedastic} -- i.e., the errors (and uncertainties) are uniform everywhere and therefore a single scalar for all samples suffices. However, in practice, most models are \emph{heteroscedastic} -- i.e., the errors around a prediction can vary for different test samples. In this paper, we propose a novel calibration strategy that estimates a specific temperature for every sample related to the unreliability or uncertainty for that sample. Unlike existing approaches, this takes the heteroscedasticity of the model into account and therefore, subsequently improves OOD detection. 
%
\section{Heteroscedastic Temperature Scaling via Neural Network Anchoring}
In this section, we first outline the proposed scoring function, followed by theoretical and empirical justification of its effectiveness in visual OOD detection. As previously stated, we are interested in learning a temperature function  $\boldsymbol\tau:\mathcal{X}\rightarrow \mathbb{R}$, defined in the image domain $\mathcal{X}$, that is used to calibrate a model's predictions. The temperature parameter for a sample $\mathrm{x}$ is denoted as $\boldsymbol\tau(\mathrm{x})$ (fixed to be the same for all classes). We refer to this process as heteroscedastic scaling, since the model predictions for any test sample can be adjusted with the learned function. Next, we discuss the proposed approach for constructing $\boldsymbol\tau$. 
\subsection{Neural network anchoring}
First, let us randomly choose a training image from the dataset $\mathcal{D}$, denoted by $\mathrm{c}$ and refer to it as an \emph{anchor}. Using this anchor $\mathrm{c}$, we define a simple coordinate transformation on the input domain as $\mathcal{E}: \mathrm{x} \rightarrow [\mathrm{c}, \mathrm{x}-\mathrm{c}]$. That is, we represent an image as a combination of the anchor, and the residual between the anchor and the image. Note that, this definition allows the use of multiple transformations (w.r.t. many anchors) to obtain predictions for a given sample $\mathrm{x}$, \textit{i.e.}, $f_{A}([\mathrm{c}_1, \mathrm{x}-\mathrm{c}_1]) = f_{A}([\mathrm{c}_2, \mathrm{x}-\mathrm{c}_2]) = \dots = f_{A}([\mathrm{c}_k, \mathrm{x}-\mathrm{c}_k])$, where $f_{A}$ refers to the model that takes the tuple $([\mathrm{c}_k, \mathrm{x}-\mathrm{c}_k])$ and predicts the target $y$. 

\paragraph{Training.} During training, for every input sample $\mathrm{x}_i$, we use one random anchor to perform the coordinate transformation, which is implemented as a simple concatenation along the channel dimension. Due to the randomness over the choice of the anchor, over the course of training each $\mathrm{x}_i$ gets combined with a large number of anchors. Since the prediction for the sample $\mathrm{x}_i$ -- regardless of the choice $\mathrm{c}$ -- is expected to be the same (label $y_i$), this enforces an implicit consistency in the predictions across different anchors. The optimization of this anchored model is similar to that of a standard network, e.g., cross entropy loss-based training.

\paragraph{Consistency via standard image augmentations.} Inspired by the recent successes of data augmentation strategies in improving model generalization, we exploit an additional consistency during training to further improve anchored models. More specifically, we modify the input to be as follows -- $\bar{\mathrm{x}} = [\mathcal{T}(\mathrm{c}), \mathrm{x} - \mathrm{c}]$, where $\mathcal{T}(.)$ refers to a pre-defined image augmentation and it returns a perturbed anchor. Intuitively, we encourage the model to learn invariances between $\mathrm{c}$ and $\mathcal{T}(\mathrm{c})$ using an asymmetry in the coordinate transformation. While different choices currently exist to implement $\mathcal{T}$ for natural images, we find that using a composition of multiple standard augmentation strategies already used in training the model (random crops, random flips, color jitter etc.) are sufficient in practice.

\begin{figure}[!htb]
 \footnotesize
 \begin{minipage}{.45\linewidth}
     \caption{\small{Pseudocode for training}}
    \begin{python}
def train_loop(trainloader,T):
    for inputs, targets in trainloader:
        A = Shuffle(inputs) 
        D = inputs-A 
        X_d = torch.cat([T(A), D],axis=1)
        y_d = model(X_d)
        loss = criterion(y_d,targets)
        ....

return
    \end{python}
\label{fig:pseudocode}
 \end{minipage}\hfill
 \begin{minipage}{.5\linewidth}
 \caption{\small{Pseudocode for inference}}
\begin{python}
def inference(inputs,anchors):
    for A in anchors:
        D = inputs-A 
        X_test = torch.cat([A, D],axis=1)
        y_test = model(X_test)
        preds.append(y_test)
    P = torch.cat(preds,0)
    H = torch.mean(P,0)
    tau = P.sigmoid().std(0).sum(1)
    ood_score = AMP(H/tau)
return H, ood_score

\end{python}
\label{fig:pseudocode2}
 \end{minipage}
 \end{figure}

\paragraph{Inference.} 
As discussed in the section, this anchoring process leads to different hypotheses for different anchor choices, so we propose to marginalize out the effect of the anchor, $\mathrm{c}$, to both obtain the predictions as well as the temperature function for performing the heteroscedastic calibration. The temperature value for a sample is estimated as the standard deviation of the predictions, obtained using multiple random anchor choices, as shown below:

\begin{align} 
\label{eqn:inference}
    H(y|\mathrm{x}) = \texttt{MEAN}\left[f_A([\mathrm{c}_k, \mathrm{x} - \mathrm{c}_k])\right]_{k=1}^K; \boldsymbol\tau(\mathrm{x}) = \mathlarger{\mathlarger{\sum}}_{\text{all classes}}\texttt{STD-DEV}\left[\sigma\bigg(f_A([\mathrm{c}_k, \mathrm{x} - \mathrm{c}_k])\bigg)\right]_{k=1}^K.
\end{align}Here, for convenience we use $\texttt{MEAN}$ and $\texttt{STD-DEV}$ to denote the mean and standard deviation over predictions obtained using $K$ different anchors during inference. Further, $H(y|\mathrm{x})$ indicates the set of logits for each of the classes. To scale the standard deviation appropriately, we compute it after passing the logits through a sigmoid activation layer. Note that, we compute the class-specific uncertainties by performing this aggregation directly using the logits obtained from the model. Essentially, we marginalize out the effect of the randomly chosen anchor to obtain the final prediction and the temperature value for a test sample.

\paragraph{Heteroscedastic temperature scaling.} Having estimated the temperature for the sample in \eqref{eqn:inference}, the heteroscedastic calibration process can be expressed as $H^c(y|\mathrm{x}) = \frac{H(y|\mathrm{x})}{\boldsymbol\tau(\mathrm{x})}$. The OOD score for this calibrated sample is given by using any pre-specified scoring function $\mathcal{S}$. In particular, we first obtain class likelihoods from the calibrated logits $P^c(y|\mathrm{x}) = \texttt{SOFTMAX}(H^c(y|\mathrm{x}))$, and compute the negative log likelihood score for OOD detection. We refer to this score as \textbf{A}nchor \textbf{M}arginalized \textbf{P}rediction score (\textbf{AMP}). Formally, the AMP scoring function can be defined as:
\begin{equation}
\label{eq:base_score}
\text{AMP}: \quad \mathcal{S}(\mathrm{x}) = -\frac{1}{N}\sum_{\text{all classes}} \log (\texttt{SOFTMAX}(H^c(y|\mathrm{x}))).
\end{equation}
For smaller datasets like CIFAR-10/100 the variance tends to be small, and to avoid scaling issues we instead use $H^c(y|\mathrm{x}) = \frac{H(y|\mathrm{x})}{1 + \exp({\boldsymbol\tau(\mathrm{x})})}$. In Figures \ref{fig:pseudocode} and \ref{fig:pseudocode2} we demonstrate pytorch-like pseudocode for training and inference.

\section{Intuition Behind \name} 
\name is effective in distinguishing OOD samples due to the proposed heteroscedastic temperature scaling strategy -- by defining the temperature value as a function of the sample, we are able to automatically adjust the scaling to be sensitive to the OOD-ness of the sample. Naturally, this works better than using a single temperature value for the entire dataset. A strong candidate for such a scaling strategy is the \emph{epistemic} uncertainty of a model for a given sample. Intuitively, scaling by uncertainty should improve OOD performance since, by definition, an OOD sample has high epistemic uncertainty, compared to an inlier, which translates to a higher temperature that scales the logits more aggressively. This leads to increased entropy in the resulting prediction probabilities. During inference, \name estimates the temperature of a sample based on the standard deviation of predictions obtained via multiple anchors. In the following section, we justify why this estimator expressed in \eqref{eqn:inference} is, in fact, related to the epistemic uncertainty.


\paragraph{} We utilize neural tangent kernel (NTK) theory \citep{jacot2018neural, arora2019fine, bietti2019inductive, lee2019wide}, as it provides a convenient framework for analyzing the effect of the modified training proposed in \name. The basic idea of NTK is that, when the width of a neural network tends to infinity and the learning rate of SGD tends to zero, the function $f(\mathrm{x};\boldsymbol{\theta})$ converges to a solution obtained by kernel regression using the NTK:
\begin{equation}
    \label{eqn:ntk}
    \mathbf{K}_{\mathrm{x}_i\mathrm{x}_j} = \mathbb{E}_{\boldsymbol{\theta}}\left\langle \pdv{f(\mathrm{x}_i,\boldsymbol{\theta})}{\boldsymbol{\theta}}, \pdv{f(\mathrm{x}_j,\boldsymbol{\theta})}{\boldsymbol{\theta}} \right\rangle.
\end{equation}When the samples $\mathrm{x}_i, \mathrm{x}_j \in \mathcal{S}^{d-1}$, i.e.,  points on the hypersphere and have unit norm, the NTK for a simple 2 layer ReLU MLP can be simplified as a dot product kernel \citep{arora2019fine,bietti2019inductive,lee2019wide}:
\begin{equation}
    \label{eqn:mlp_ntk}
\mathbf{K}_{\mathrm{x}_i\mathrm{x}_j} =     h_{\text{NTK}}(\mathrm{x}_i^\top \mathrm{x}_j) = \frac{1}{2\pi}\mathrm{x}_i^\top \mathrm{x}_j(\pi - \cos^{-1}(\mathrm{x}_i^\top \mathrm{x}_j)).
\end{equation}
Let us also consider the prediction on a test sample $\mathrm{x}_t$ in the limit as the inner layer widths grow to infinity. It has been shown that (c.f. \citep{lee2019wide,bietti2019inductive}): 
\begin{equation}
\label{eqn:ntk_limit}
\centering
f_\infty(\mathrm{x}_t) = f_0(\mathrm{x}_t) - \mathbf{K}_{\mathrm{x}_t\mathbf{X}}\mathbf{K}_{\mathbf{X} \mathbf{X}}^{-1}(f_0(\mathbf{X})-\mathbf{Y}),
\end{equation}where $f_0$ is the network with the initial random weights, $\theta_0$, and $\mathbf{X}$ is the matrix of all training data samples.
\paragraph{NTK with neural network anchoring.}
Recall, the anchoring process involves transforming the input into a tuple: $[\mathrm{c},\mathrm{x}-\mathrm{c}]$, for a randomly chosen anchor $\mathrm{c}$. In the following, we examine the impact of anchoring on the NTK. 
%
%
Without loss of generality, we assume $[\mathrm{c},\mathrm{x}_i-\mathrm{c}]$ and $[\mathrm{c},\mathrm{x}_j-\mathrm{c}]$ are unit norm, so we can simplify $h_{\text{NTK}}([\mathrm{c},\mathrm{x}_i-\mathrm{c}]^\top [\mathrm{c},\mathrm{x}_j-\mathrm{c}])$. Further, we use a Taylor series approximation for $\cos^{-1}$: $\cos^{-1}(\mathrm{u}-\mathrm{c}) \approx \cos^{-1}(\mathrm{u}) + \frac{\mathrm{c}}{\sqrt{1-(\mathrm{u}-\mathrm{c})^2}}$. 
\begin{figure}[!htb]
    \centering
    \includegraphics[width=.95\linewidth,trim=2cm 0cm 2cm 2cm,clip]{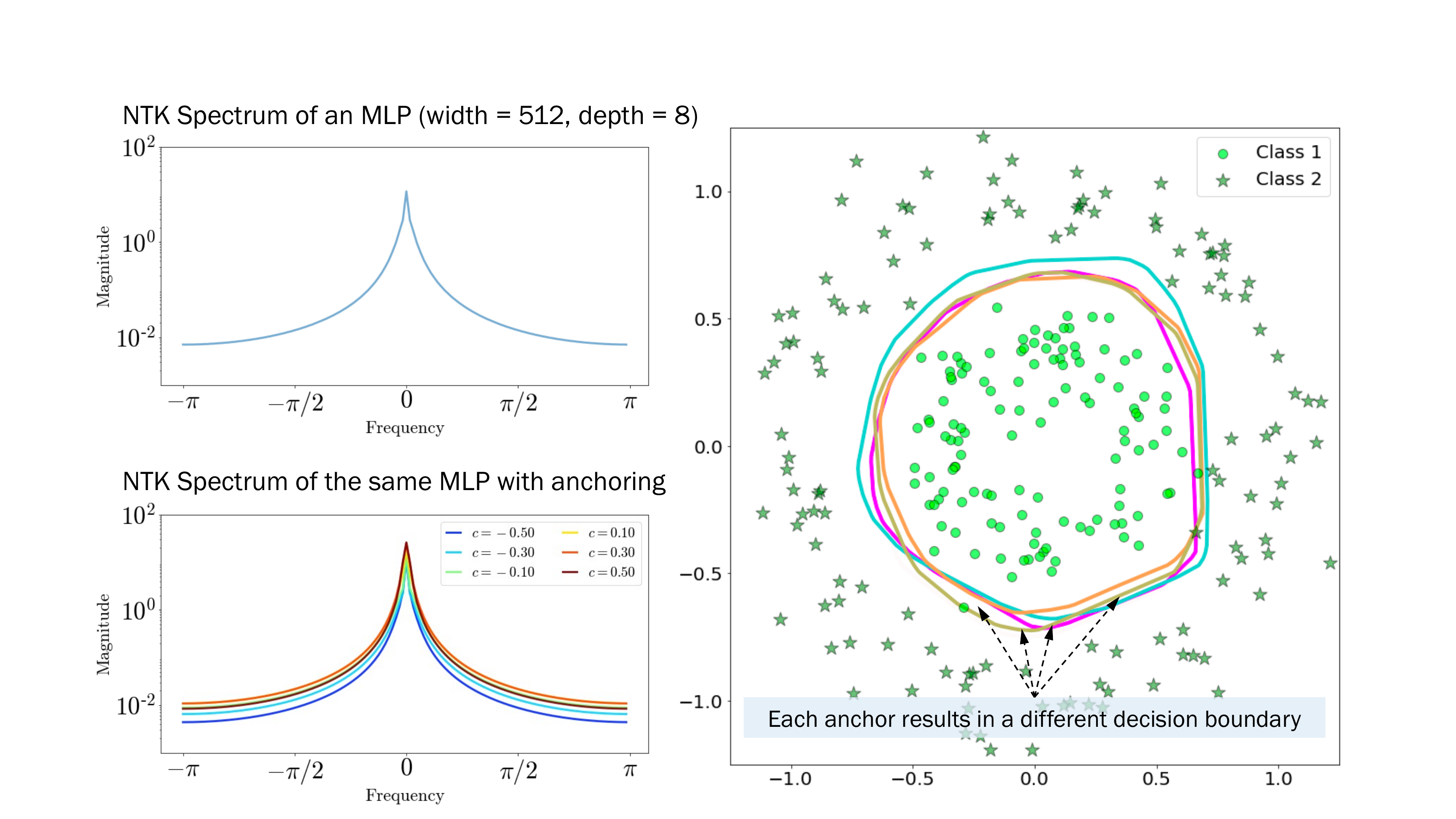}
    \caption{\small{NTK Spectra for a vanilla model and one with neural network anchoring (left). In the simple classification setup on the right, we demonstrate the effect of different anchors on the classifier's predictions. We observe that each anchor produces a slightly different NTK (due to \eqref{eqn:shifted_ntk}), resulting in meaningful inconsistencies in the classifier's predictions.}}
    \label{fig:ntk_illus}
\end{figure}

Expanding $[\mathrm{c},\mathrm{x}_i-\mathrm{c}]^\top [\mathrm{c},\mathrm{x}_j-\mathrm{c}]=\mathrm{x}_i^\top \mathrm{x}_j - \mathrm{c}^\top(\mathrm{x}_i + \mathrm{x}_j - 2\mathrm{c})$ and letting $\mathrm{v} = (\mathrm{x}_i + \mathrm{x}_j - 2\mathrm{c})$, we obtain the expression for $h_{\text{NTK}}$ under a shifted domain (from \eqref{eqn:mlp_ntk}) as follows:
\begin{flalign}
\label{eqn:shifted_ntk}
    & \mathbf{K}_{\text{anc}} =  \frac{1}{2\pi}(\mathrm{x}_i^\top \mathrm{x}_j - \mathrm{c}^\top \mathrm{v})(\pi - \cos^{-1}(\mathrm{x}_i^\top \mathrm{x}_j - \mathrm{c}^\top \mathrm{v})) \notag &\\
    & \approx \frac{1}{2\pi}\mathrm{x}_i^\top \mathrm{x}_j(\pi - \cos^{-1}(\mathrm{x}_i^\top \mathrm{x}_j)) - \frac{1}{2\pi}\mathrm{c}^\top \mathrm{v}(\pi - \cos^{-1}(\mathrm{x}_i^\top \mathrm{x}_j)) - \frac{\mathrm{c}(\mathrm{x}_i^\top \mathrm{x}_j - \mathrm{c}^\top \mathrm{v})}{2\pi \sqrt{1 - (\mathrm{x}_i^\top \mathrm{x}_j - \mathrm{c}^\top \mathrm{v})^2}}\notag&\\
    & = \mathbf{K}_{\mathrm{x}_i\mathrm{x}_j} - \Gamma_{\mathrm{x}_i, \mathrm{x}_j, \mathrm{c}},
\end{flalign}
where we combine all terms dependent on $c$ into $\Gamma_{\mathrm{x}_i, \mathrm{x}_j, \mathrm{c}}$, which also behaves as a dot product kernel. We can see that combining this \emph{stochastic} NTK (in $\mathrm{c}$) into \eqref{eqn:ntk_limit}, the prediction on a test sample is correspondingly stochastic under a fixed initialization $\theta_0$. In other words, neural network anchoring effectively perturbs the hypothesis space of the neural network resulting in an ensembling-like behavior, that effectively produces different predictions, conditioned on the anchor choice. We illustrate the process in Figure \ref{fig:anchoring_illus}, and compute the NTK spectrum (as done in \citeauthor{tancik2020fourier}) along with a demonstrative toy classification problem in \cref{fig:ntk_illus}. Our training objective (with randomly chosen anchors) forces the predictions on a sample across anchors to be the same for in distribution samples, and therefore the disagreement (measured by sum of standard deviation, \eqref{eqn:inference}) is larger when the sample is OOD.  This is similar, in principle, to deep ensembles \citep{ovadia2019}, that measure prediction variance by training multiple models with different random initializations $\boldsymbol{\theta}_0$.



\section{Experiments and Results}
In this section we evaluate the proposed anchoring-based OOD detection on a wide variety of OOD benchmarks, with several different model architectures, and OOD settings.

\paragraph{Setup.} Our modification to any standard neural network architecture is minimal -- we only change the first convolutional layer to accept a $6$ channel input (3 channels for anchor and residual each) instead of the original $3$ channels. We use standard hyper-parameters to train all our models. For our experiments with CIFAR10/100 \citep{krizhevsky2009learning} datasets, we trained a wide ResNet (WRN) model for 200 epochs, with an initial learning rate of 0.1, and a decay of 0.2 at $60$, $120$, and $160$ epochs. In addition to WRN, we also experimented with ResNet-18 and ResNet-34 architectures. For our ImageNet \citep{russakovsky2015imagenet} experiments we trained a ResNet-50 model with an initial learning rate of 0.1 and a decay of 0.1 every 30 epochs. Note that, we used the same normalization scheme for pre-processing both in- and out- distribution data sets (at train and test times). As discussed in the previous section, for our method, we used a single randomly chosen anchor for every input image in a mini-batch. From our extensive empirical studies, we observed no significant difference in the top-1 accuracy of the anchored models on any of the benchmarks. For example, with the WRN-40-2 on CIFAR-10, we obtained an accuracy of $95.1$ on average, and on CIFAR-100 a top-1 accuracy of $76.1$. On ImageNet, our ResNet-50 model trained for $120$ epochs resulted in a top-1 accuracy of $76.0$. More details of the benchmarks and the exact hyper-parameter settings are provided in the supplement.

\paragraph{Baselines.} We performed comparisons with several widely-adopted OOD detection approaches: (a) Maximum Softmax Probability (MSP); (b) ODIN \citep{odin}; (c) Energy \citep{energyood}; (d) Gram Matrices GM \citep{sastry2020detecting} that uses latent space deviation to detect OOD, but has been shown to perform better than scoring functions such as Mahalanobis detection \citep{mahalanobis} without requiring exposure to outlier data. We also evaluated our approach against uncertainty-based OOD detectors for the ImageNet experiments, namely MC-dropout \citep{gal2016dropout}, deep ensembles \citep{lakshminarayanan2016simple} and techniques that require post-hoc calibration on a validation set like temperature scaling \citep{guo2017calibration}, SVI \citep{blundell2015weight}, and AVuC \citep{krishnan2020improving}. 

\noindent \textbf{Metrics:} We used $4$ standard metrics to evaluate OOD detection performance across our benchmarks -- (a) \textbf{FPR95}: False positive rate of examples from the OOD set when the true positive rate (TPR) of the in-distribution is set at 95\%; (b) \textbf{AUROC}: Area under the receiver operating characteristic curve; (3) \textbf{AUPR}: Area under the precision-recall curve for both In/Out sets depending on which one is considered positive; (4) \textbf{DTACC}: Detection accuracy measures the maximum possible OOD detection accuracy across all thresholds as proposed in \citep{mahalanobis}.

\begin{table*}[!htbp]
\centering
\resizebox{\linewidth}{!}{
\begin{tabular}{p{.05in}llll}
\toprule
\multirow{2}{*}{\begin{tabular}[c]{@{}c@{}}\end{tabular}}      & \multicolumn{1}{c}{\multirow{2}{*}{OOD}} & \multicolumn{1}{c}{FPR95 $\downarrow$}         & \multicolumn{1}{c}{AUROC $\uparrow$}                   & \multicolumn{1}{c}{AUPR $\uparrow$}           \\
\cmidrule{3-5}
                                                                               & \multicolumn{1}{c}{}                     & \multicolumn{3}{c}{MSP / ODIN / Energy / \name (ours)}                                                                                 \\
\midrule

\multirow{7}{*}{\begin{tabular}[c]{@{}c@{}}\rotatebox[origin=c]{90}{\textbf{CIFAR-10}}\end{tabular}}  & iSUN         & 56.03 / 32.05 / 33.68 / \textbf{16.59}         & 89.83 / 93.50 / 92.62 / \textbf{97.16}   & 97.74 / \textbf{98.54} / 98.27 / 97.83 \\
                                                                                & LSUN (R)      & 52.15 / 26.62 / 27.58 / \textbf{13.73}         & 91.37 / 94.57 / 94.24 / \textbf{97.77}   & 98.12 / \textbf{98.77} / 98.67 / 97.77 \\
                                                                                & LSUN (C)      & 30.80 / 15.52 /~~~8.26 /~~~\textbf{1.50}         & 95.65 / 97.04 / 98.35 / \textbf{99.55}   & 99.13 / 99.33 /  \textbf{99.66} / 99.57 \\
                                                                                & Places365     & 59.48 / 57.40 / 40.14 / \textbf{19.89}         & 88.20 / 84.49 / 89.89 / \textbf{95.79}   & 97.10 / 95.82 /  \textbf{97.30} / 87.28 \\
                                                                                & Texture       & 59.28 / 49.12 / 52.79 / \textbf{35.43}         & 88.50 / 84.97 / 85.22 / \textbf{93.61}   & \textbf{97.16} / 95.28 / 95.41 / 96.14  \\
                                                                                & SVHN          & 48.49 / 33.55 / 35.59 / ~~\textbf{5.19}         & 91.89 / 91.96 / 90.96 / \textbf{98.10}   & \textbf{98.27} / 98.00 /  97.64 / 97.51 \\
\cmidrule{2-5}
& \textit{Average} & 42.71 / 35.71 / 33.01	/ \textbf{15.39} &
90.91 / 91.01 / 91.88 / \textbf{96.99} & \textbf{97.91} / 97.62 / 97.82 / 96.02 \\
\cmidrule{2-5}
\multirow{7}{*}{\begin{tabular}[c]{@{}c@{}}\rotatebox[origin=c]{90}{{\textbf{CIFAR-100}}}\end{tabular}}& iSUN          & 82.80 / 68.51 / 81.10 / \textbf{67.15}         & 75.46 / 82.69 / 78.91 / \textbf{83.79}  & 94.06 / \textbf{95.80} /94.91 / 85.84   \\
                                                                              & LSUN (R)        & 82.42 / 71.96 / 79.47 / \textbf{61.73}         & 75.38 / 81.82 / 79.23 / \textbf{85.64}  & 94.06 / \textbf{95.65} / 94.96 / 84.30   \\
                                                                              & LSUN (C)        & 66.54 / 55.55 / 35.32 / ~~\textbf{4.16}         & 83.79 / 87.73 / 93.53 / \textbf{99.18}  & 96.35 / 97.22 / 98.62 / \textbf{99.19}   \\
                                                                              & Places365       & 82.84 / 87.88 / 80.56 / \textbf{65.18}         & 73.78 / 71.63 / 75.44 / \textbf{85.78}  & 93.29 / 92.56 / \textbf{93.45} / 69.65    \\
                                                                              & Texture         & 83.29 / \textbf{79.27} / 79.41 / 81.81         & 73.34 / 73.45 / \textbf{76.28} / 71.36  & 92.89 / 92.75 / \textbf{93.63} / 79.72   \\
                                                                              & SVHN            & 84.59 / 84.66 / 85.82 / \textbf{12.57}         & 71.44 / 67.26 / 73.99 / \textbf{97.85}  & 92.93 / 91.38 / 93.65 / \textbf{95.25} \\

\cmidrule{2-5}
& \textit{Average} & 80.41 / 74.64 / 73.61 / \textbf{48.77} &
75.53 /	77.43 / 79.56 /	\textbf{87.27} & 93.93 / 94.23 / \textbf{94.87} / 85.66 \\

\bottomrule
\end{tabular}
}
\caption{\small{\textbf{OOD Detection with WideResNet-40-2:} We compare with a range of different OOD detection methods on the commonly used OOD benchmark. We see a consistent improvement in performance when using \name, over existing baselines -- across both CIFAR-10/100 models. \name is particularly good in suppressing false positives as reflected by the FPR95 metric.}}
\label{tab:ood_benchmark1}
\end{table*}
\subsection{Performance on OOD Benchmarks} 
\vspace{-5pt}
We begin by evaluating \name on a commonly used OOD detection benchmark first introduced by Liang \textit{et al.} \citep{odin}. In this experiment, predictive models were trained with CIFAR-10 or CIFAR-100 as the in-distribution, and then used to detect out-of-distribution data chosen from one of the six datasets (details in the supplement). We follow the experimental protocol from~\citep{energyood}, and used the WideResNet architecture WRN-40-2 \citep{zagoruyko2016wide}. In \Cref{tab:ood_benchmark1}, we show results across all the six datasets for both CIFAR-10/100, using the three metrics. On average, we find that \name significantly improves on the challenging FPR95 metric (nearly ~$50\%$ lower than the next best), while also providing substantial gains in the AUROC metric. We notice that, these gains persist even with CIFAR-100, which is known to be a much harder setting (as reflected by the higher false positive rates). We do not assume access to additional OOD data (for model finetuning), unlike both energy- and ODIN-based detection that perform similarly, while our approach significantly improves upon them ($\approx 25\%$ drop in the FPR metric). 
\begin{table}[!t]
     \resizebox{.95\linewidth}{!}{
    \begin{minipage}{0.6\linewidth}
        \centering
    \footnotesize
    \begin{tabular}{lllll}
    
    \toprule
       & Method & FPR95 $\downarrow$ & AUROC $\uparrow$ & DTACC $\uparrow$ \\
        \midrule
      \multirow{6}{*}{\begin{tabular}[c]{@{}c@{}}\rotatebox[origin=c]{90}{\textbf{CIFAR-10}}\end{tabular}}   
        & MSP & 55.12 & 90.37 &84.65 \\
        & ODIN  & 33.40 &92.70 &  85.28 \\
        & Energy& 28.89 & 94.47 & 88.58\\
        & \small{Mahal.*} &  14.8 & \textbf{97.33} & \textbf{93.15}\\
        & GM & 14.83 & 96.33 & 92.14 \\
        & \name (ours) & \textbf{12.33} & \textbf{97.20}  & 92.84 \\
        
     \cmidrule{2-5}
      \multirow{6}{*}{\begin{tabular}[c]{@{}c@{}}\rotatebox[origin=c]{90}{\textbf{CIFAR-100}}\end{tabular}}   
      & MSP & 80.22 & 77.22 & 70.95 \\
        & ODIN & 60.82 & 85.41 & 78.61 \\
        & Energy& 70.96& 82.44& 75.67 \\
        & \small{Mahal.*} &  \textbf{24.36} & \textbf{94.07} & \textbf{88.51}\\
        & GM & 49.87 & 89.96 & 83.50 \\
        & \name (ours) & 49.70 & 89.97  & 82.50 \\    
        \bottomrule
    \end{tabular}
    \caption{\small{\textbf{OOD detection performance with ResNet-34:}. Averaged across 7 datasets following \cite{sastry2020detecting}. Here, `Mahal.' indicates the Mahalanobis score, which uses additional outlier data during training.}}
    \label{tab:resnet34_ood}
    \end{minipage}%
    \qquad 
    \begin{minipage}{.4\linewidth \relax}
    \footnotesize
    \centering
    \begin{tabular}{l|ll}
    \toprule
    \multirow{2}{*}{} & \multirow{2}{*}{Method} & ResNet-34   \\
    && \multicolumn{1}{c}{\tiny{FPR95 $\downarrow$ / AUROC $\uparrow$}} \\
    \midrule
    \multirow{4}{*}{\rotatebox[origin=c]{90}{\textbf{\tiny{C10 $\rightarrow$ C100}}}}
                                                    & ODIN     & 58.0 / 88.2 \\
                                                    & Energy   & 47.5 / 88.4 \\
    \multicolumn{1}{l|}{}                           & GM     & 59.8 / 83.6 \\
    \multicolumn{1}{l|}{}                           & Mahal.$^*$     & 58.4 / 88.2 \\
    \multicolumn{1}{l|}{}                           & \name (ours)  & \textbf{43.5 / 90.2} \\ 
    \cmidrule{2-3}
    \multicolumn{1}{l|}{\multirow{4}{*}{\rotatebox[origin=c]{90}{\textbf{\tiny{C100 $\rightarrow$ C10}}}}} 
                                                    & ODIN     & 81.3 / 77.2 \\
                                                    & Energy    & 80.9 / 77.0 \\
    \multicolumn{1}{l|}{}                           & GM     & 83.1 / 74.5 \\
    \multicolumn{1}{l|}{}                           & Mahal.$^*$    & \textbf{79.8} / 77.5 \\
    \multicolumn{1}{l|}{}                           & \name   & 82.5 / \textbf{79.9} \\ 
    \bottomrule
    \end{tabular}
    \caption{\small{\name is effective on near OOD detection with CIFAR-10$\leftrightarrow$100 respectively.}}
    \label{tab:near-ood}
    
    \end{minipage}%
    }
    \end{table}

Next, in \Cref{tab:resnet34_ood}, we evaluate on the benchmark introduced in \citep{sastry2020detecting} using ResNet-34. This consists of $7$ different OOD datasets -- iSUN, LSUN (R), LSUN (C), TinyImageNet (R), TinyImageNet (C), CIFAR10/100, SVHN. We observe that \name comes second to Mahalanobis on CIFAR-100 while matching it on CIFAR-10, even though we do not use outlier data for fine-tuning. In order to make a fair comparison on the same OOD samples for all datasets, we use GM without the validation data (and corresponding normalization) -- the performance improvements were consistently observed in most cases. Detailed results for this benchmark are available in the supplement.
\begin{table}
\centering
\resizebox{0.9\linewidth}{!}{
\begin{tabular}{p{3in}|p{0.6in}p{0.6in}c}
	\toprule
    Method & AUROC~$\uparrow$& DTACC~$\uparrow$& AUPR-in/out~$\uparrow$\\ 
    \midrule
	ResNet-50 \citep{he2016deep} & 93.36 & 86.08 & 92.82 / 93.71 \\
	Temp-Scal \citep{guo2017calibration} & 93.71 & 86.47& 93.21 / 94.01 \\
	Deep Ens \citep{lakshminarayanan2016simple} & 95.49 & 88.82 & 95.31 / 95.64\\
	MCD \citep{gal2016dropout} & 96.38 & 89.98 & 96.16 / 96.67\\
	SVI \citep{blundell2015weight} & 96.40 & 90.03 & 95.97 / 96.83\\
	SVI-AvUC \citep{krishnan2020improving} & 97.60 & 92.07 & 97.39 / 97.85\\
	\midrule
	\name (ours) & \textbf{99.07} & \textbf{95.14} &  \textbf{99.10} /  \textbf{99.08} \\
	\bottomrule
\end{tabular}
}
\caption{\small{\textbf{UQ-based detection of ImageNet-C with ResNet-50:} We evaluate \name on the benchmark introduced by \citep{krishnan2020improving} where we use Gaussian Blur of level 5 intensity as the OOD, and the clean ImageNet validation as the ID. For comparisons with OOD methods on this benchmark, refer to \Cref{fig:imagenet_near_ood}.}}
\label{tab:Imagenet}
\end{table}


\noindent \textbf{Uncertainty based ImageNet-OOD detection.} The prediction variance from \name can be interpreted as a measure of unreliability in the model's predictions. Hence, a natural evaluation is to leverage this unreliability estimate as a score for OOD detection, since it is expected to be high as we move away from the original data manifold (\textit{i.e.}, OOD), while being low for in-distribution data. We analyzed this performance on ImageNet-C data~\citep{hendrycks2018benchmarking} using a ResNet-50 architecture trained on ImageNet-1K, as shown in \Cref{tab:Imagenet}. Specifically, we used the OOD dataset obtained via Gaussian blur corruption at intensity $5$. We find that, by leveraging our unreliability score to perform adaptive temperature scaling, \name is highly effective for OOD detection. In fact, this improves over several state-of-the-art uncertainty estimators on this benchmark.
\begin{figure*}[!htb]
    \centering
        \subfigure[\small{Densities of scores between in and out distribution ImageNet-C.}]{\includegraphics[width=0.36\linewidth]{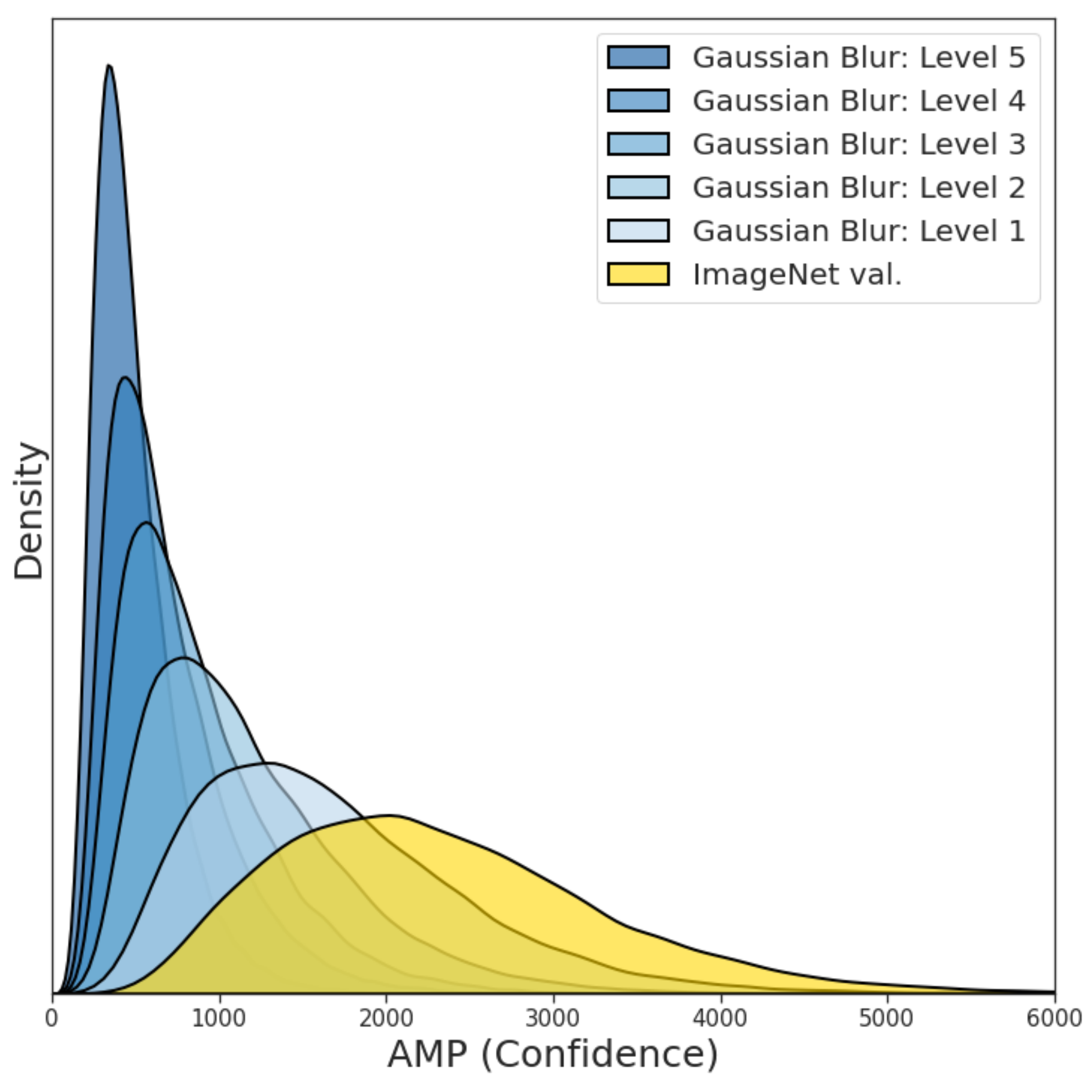}
        \label{fig:imagenet_density}}
        \qquad
        \subfigure[\small{AUROC $(\uparrow)$ across corruption levels for \name}]{\includegraphics[width=0.5\linewidth]{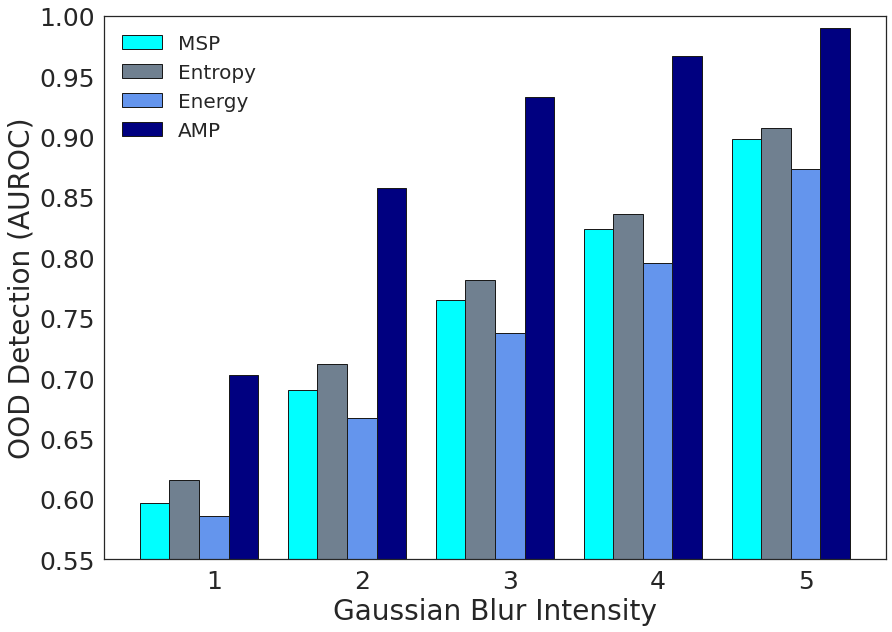}
        \label{fig:imagenet_near_ood}}
    \caption{\small{\textbf{From near to far OOD:} \name outperforms baselines consistently at all the intensity levels (\cref{fig:imagenet_near_ood}), and produces meaningful scores (\cref{fig:imagenet_density}) when moving from near (level 1) to far (level 5) OOD sets.}}
    \label{fig:imagenet-ood}
    \vspace{-15pt}
\end{figure*}


\noindent \textbf{{Near OOD detection.}} Here, we consider the more challenging ``near OOD'' detection task that seeks to separate CIFAR-10 and CIFAR-100 datasets, by using a model trained on one of them while treating the other as the OOD set. Since both are drawn from the same image distribution but have mutually exclusive classes, this is extremely challenging and often causes OOD detectors to fail. We measure OOD performance using both FPR and AUROC as shown in \Cref{tab:near-ood}. We see that methods like Mahalanobis detector \citep{mahalanobis}, Gram Matrices \citep{sastry2020detecting}, which are otherwise very competitive tends to fail in this task -- whereas \name is able to separate them better, indicating its effectiveness for both near and far OOD tasks.

\noindent We also study how \name performs as the OOD set is artificially made to go farther from the ImageNet validation set using the Gaussian blur corruption of varying intensity. We expect that an effective scoring mechanism must reflect this well in its scores. We demonstrate this in \cref{fig:imagenet_density} that shows the kernel density plot of \name scores for both in- and out-distribution datasets across 5 intensity levels. Since the score is designed so as to be higher for the in-distribution set, we find a gradual and meaningful transition between the clean in-distribution set to the farthest OOD set (tight in the low confidence area). This is reflected in the OOD detection performance as measured by AUROC shown in \cref{fig:imagenet_near_ood}, where all three variants of scoring functions with \name outperform competing approaches consistently across different corruption levels. 
\begin{table*}[!htb]
\small
\setlength{\tabcolsep}{6pt}
\centering
\resizebox{.9\linewidth}{!}{
\begin{tabular}{cl|cccc}
\toprule
\multirow{2}{*}{In-distribution} 
&\multirow{2}{*}{Method} 
&\multirow{2}{*}{\begin{tabular}[c]{@{}c@{}}\small{Needs OOD}\\ \small{Exposure?}\end{tabular}} 
&\multirow{2}{*}{FPR95~$\downarrow$} 
&\multirow{2}{*}{AUROC~$\uparrow$} 
&\multirow{2}{*}{AUPR(In/Out)~$\uparrow$}\\
\\\midrule
\multirow{5}{*}{\begin{tabular}[c]{@{}c@{}}CIFAR-10\\ (ResNet-18)\end{tabular}}
& ODIN~\citep{odin}         & \xmark & 52.00 & 82.00 & 73.13~/~85.12 \\ 
& Energy~\citep{energyood}      & \xmark  & 50.03 &	83.83 &	77.15~/~85.11 \\
& OE~\citep{hendrycks18oe}  & \cmark & 50.53	& 88.93	& 87.55~/~87.83	\\
& MCD~\citep{mcd}           & \cmark & 73.02&	83.89&	83.39~/~80.53\\
& UDG \citep{scood}   & \cmark & \textbf{36.22} & \textbf{93.78} & \textbf{93.61}~/~\textbf{92.61} \\ 
\cmidrule{2-6}
& \name & \xmark  & \textbf{36.82} & 92.40 & 91.23~/~90.91 \\ 
\midrule
\multirow{5}{*}{\begin{tabular}[c]{@{}c@{}}CIFAR-100\\ (ResNet-18)\end{tabular}}
& ODIN~\citep{odin}        & \xmark  &81.89	&77.98	&78.54~/~72.56	\\
& Energy~\citep{energyood}     & \xmark &81.66	&79.31	&80.54~/~72.82	\\
& OE~\citep{hendrycks18oe} & \cmark &80.06	&78.46	&80.22~/~71.83	\\
& MCD~\citep{mcd}          & \cmark & 85.14	&74.82	&75.93~/~69.14  \\
& UDG \citep{scood}     & \cmark & 75.45 & 79.63  & 80.69~/~74.10\\ 
\cmidrule{2-6}
& \name & \xmark   & \textbf{70.34} & \textbf{82.22} & \textbf{84.14~/~76.20} \\ 
\bottomrule
\end{tabular}
}

\caption{\small{\textbf{SCOOD benchmark with ResNet-18.} \name outperforms even the best performing techniques on the recent semantically coherent OOD bencmark \citep{scood} on both CIFAR-10 and CIFAR-100 inspite of not requiring outlier exposure. The methods OE~\citep{hendrycks18oe}, MCD~\citep{mcd}, and UDG use Tiny-ImageNet during training.}}
\vspace{-20pt}
\label{tab:scood}
\end{table*}
\subsection{Semantically Coherent OOD (SCOOD)}

The current suite of OOD detection benchmarks rely on separating the dataset on which a model is trained from another dataset marked as OOD, for example, CIFAR-SVHN. However, considering that dataset specific biases are prominent, it is likely that many of the OOD detection algorithms tend to overemphasize dataset-specific \emph{noise} in metrics such as FPR or AUROC. Further, it is likely there are factors other than semantic content that are contributing to very high OOD performance. To address this, we consider the recently proposed SCOOD (semantically coherent OOD detection) benchmark \citep{scood}, which effectively re-samples the in- and out-distribution datasets such that the only truly distinguishing factor between the two is the semantic content. This also includes transferring semantically similar images from the OOD dataset into the in-distribution set (e.g., cats from TinyImageNet into CIFAR-100), and getting rid of resizing artifacts so that the OOD performance reflects the true performance.

\noindent In \Cref{tab:scood} we report average OOD detection performance with \name on the SCOOD benchmark, which is comprised of $6$ different resampled/resized datasets -- Texture, SVHN, CIFAR-10/100, Tiny-ImageNet, LSUN, Places365. We trained ResNet-18 models on modified CIFAR-10/100 training sets provided by SCOOD \citep{scood} and test on their custom OOD sets for fair comparison. We compare the different methods on FPR95, AUROC, and AUPR (In/Out) metrics. The current best performing approach on SCOOD, UDG \citep{scood} additionally also uses outlier data from TinyImageNet for training, similar to other approaches like OE. We find that \name is the best performing method on CIFAR-100 across all the metrics, while on CIFAR-10 \name performs comparably to UDG in terms of FPR95, while being a close second on the other metrics -- in spite of not having any access to outlier data, while being significantly better than other comparable baselines. 
\begin{table*}[!htb]
\centering
\footnotesize
\resizebox{0.95\linewidth}{!}{
\begin{tabular}{l|l|llll|l}
\hline
\multirow{2}{*}{In Distribution} & \multirow{2}{*}{Method} & \multicolumn{4}{c|}{\textbf{Pillow Resizing from original LSUN} } &  \multirow{2}{*}{Average}\\
& \multicolumn{1}{c|}{}   &  \texttt{nearest}* & \texttt{bilinear} & \texttt{bicubic} & \multicolumn{1}{l|}{\texttt{lanczos}} &  \\ \cmidrule{1-7}
 \multirow{4}{*}{\begin{tabular}[c]{@{}c@{}}CIFAR-10\\(ResNet-34)\end{tabular}} 
                            & MSP   &  41.5 / 94.0 & 47.8 / 91.6 & 45.5 / 92.2 & \multicolumn{1}{l|}{45.3 / 92.4} & 46.9 / 92.2 \\ 
                            & Energy  &  28.6 / 98.4 & 34.5 / 92.9 & 33.0 / 93.4 & \multicolumn{1}{l|}{32.0 / 93.8} & 30.4 / 94.1\\ 
                            & GM  & \textbf{~~1.8 / 99.2}& 46.2 / 90.7 & 49.0 / 90.6 & \multicolumn{1}{l|}{46.3 / 91.3} & 25.6 / 94.9 \\
                            \cmidrule{2-7}
                            & \name  & ~~7.1 / 98.4  & \textbf{13.0 / 97.4} & \textbf{13.9 / 97.2} & \textbf{14.3 / 97.2} & \textbf{~~9.6 / 98.1} \\
\midrule
 \multirow{4}{*}{\begin{tabular}[c]{@{}c@{}}CIFAR-100\\(ResNet-34)\end{tabular}} 
                            & MSP   &  69.0 / 83.8          & 80.0 / 79.6          & 84.0 / 79.9          & \multicolumn{1}{l|}{84.0 / 80.5} &  79.9 / 79.2 \\
                            & Energy   & 52.4 / 89.3          & 84.9 / 74.9          & 83.3 / 75.7          & \multicolumn{1}{l|}{81.4 / 76.8} & 73.9 / 79.6 \\
                            & GM    & \textbf{38.7 / 94.4} & 78.1 / 79.0 & 77.6 / 80.0 & \multicolumn{1}{l|}{76.0 / 81.5} & 60.9 / 86.5 \\
                            \cmidrule{2-7}
                            & \name & 51.1 / 90.5 & \textbf{75.4 / 81.2} & \textbf{74.0 / 81.7} & \textbf{71.8 / 82.7} & \textbf{58.3 / 86.9} \\
\bottomrule
\end{tabular}}
\hspace{\fill} \footnotesize{$^*$aliasing artifacts are likely ~~~~ Performance measures are FPR95 $\downarrow$ / AUROC $\uparrow$}
\caption{\small{\textbf{Resizing methods vs OOD benchmarks:} We study the effect of resizing on OOD performance. Here we use ResNet-34 trained on CIFAR-100 as our ``in'' dataset, to detect 10K LSUN test images. These are resized using different interpolation algorithms from the Pillow package. We evaluate the performance of different OOD detection algorithms on these various datasets, and observe that \name consistently performs well across all the variants of LSUN.}} 
\label{tab:resizing_ood}
\vspace{-10pt}
\end{table*}

\subsection{Robustness to Resizing Artifacts}
As stated previously, OOD detection benchmarks can be hard to interpret when OOD datasets are laden with their dataset-specific noise/biases, often reflecting in overly optimistic OOD performance. A key source for such kinds of noise is artifacts obtained from resizing -- since most datasets have different sized images, they are resized typically using a nearest neighbor interpolation algorithm before running any OOD detection algorithm. The issue with resizing packages like OpenCV, or native resizing in Pytorch or Tensorflow was recently studied in detail in~\citep{parmar2021buggy} with its impact on FID scores for generative models -- they concluded that the Pillow resizing with a \texttt{bicubic} interpolation scheme was the most reliable and gets rid of aliasing artifacts the best. Designing specific resizing frameworks has also been shown to have an impact on classification accuracy \citep{talebi2021learning}. 

\paragraph{}This issue is further exaggerated in the current OOD setup relying on CIFAR-10/100 because the images are very small ($32\times 32$) when compared to other datasets. Specifically for OOD, this was pointed out in the SCOOD benchmark \citep{scood}, as being one of the reasons to re-design these benchmarks from scratch, however there is no ablation on how resizing alone actually affects performance. Here, we evaluate how some of the best OOD detection algorithms perform on the same in-out distribution experiment -- CIFAR-10/100 $\rightarrow$ LSUN. Since the original LSUN images are of a much larger size, we resize them using the Pillow resizing library using different types of interpolations. 

We report the results of this study in \Cref{tab:resizing_ood}. The first striking observation is the amount of variance in OOD performance across the different LSUN variants -- across the baselines, and metrics we used. In particular -- the case of \texttt{nearest} interpolation is the most similar to the LSUN (R) benchmark , and it is expected to introduce the most amount of aliasing artifacts, and existing approach produce unusually low FPR scores. However, the results the more sophisticated interpolation methods are expected to be more indicative of the true OOD performance, since they are not prone to aliasing. Interestingly, we note that \name performs the best on all of these cases. Our trend across datasets is similar to energy-based OOD \citep{energyood} and on average, \name significantly outperforms the other baselines on both the metrics considered.

\subsection{Ablation Studies} 
Finally, we study the two important aspects of \name here and their impact on OOD performance using the WRN-40-2 on the CIFAR-10/SVHN benchmark. In \Cref{tab:ablations}, we report FPR95 as our metric of choice since its the most sensitive and reflective of the performance. We ablate on two factors -- the type of transformation used during training to anchor the neural network, and the number of anchors needed during inference time. We see that simple functions like Gaussian blur or color jitter drastically improve the performance. In all our experiments, we used a combination of all the five corruptions, with 5 anchors. 
\begin{wraptable}{r}{7.5cm}
\small
\centering
\resizebox{\linewidth}{!}{
\begin{tabular}{l|cccc}
\toprule
\diagbox[width=10em]{Corruption}{$\#$ Anchors} & \textbf{2} & \textbf{5} & \textbf{10} & \textbf{20} \\
\midrule
None (trivial)           & 50.71      & 50.73      & 50.85       & 50.89       \\
\midrule
\begin{tabular}[c]{@{}l@{}}+ \texttt{ColorJitter}\\ + \texttt{GaussianBlur}\end{tabular} & ~~6.53       & ~~6.31       & ~~6.36        & ~~6.28        \\
\midrule
\begin{tabular}[c]{@{}l@{}}+ \texttt{HorizontalFlip}\\ + \texttt{Grayscale}\end{tabular}  & ~~9.96       & ~~9.85       & 10.11       & ~~9.89        \\
\midrule
+ \texttt{ResizedCrop}                                                           & ~~5.32       & ~~5.19       & ~~5.35        & ~~5.22\\       
\bottomrule
\end{tabular}
}
\caption{\small{\textbf{Ablation studies} demonstrating FPR95 for CIFAR-10-SVHN OOD experiment using types of transformations and varying number of anchors. All these transformations are applied at random every time they are executed.}}
\vspace{-30pt}
\label{tab:ablations}
\end{wraptable} 
As can be seen, there is not a significant difference in performance while increasing the number of anchors, but a big boost in using consistency training using any of the corruption functions. 

\section{Discussion}
\vspace{-5pt}
In this paper, we introduced \emph{anchoring} as a strategy to achieve effective heterscedastic temperature scaling for state-of-the-art OOD detection on a large suite of benchmarks including near and semantically coherent-OOD problems. Using NTK theory, we show that our temperature estimates are closely linked to epistemic uncertainty of the classifier, explaining it superior performance. We also introduced a new benchmark that evaluates the robustness of OOD detection methods against resizing artifacts. Anchoring is a powerful new mechanism to estimate confidence or reliability of a model, and shows promise well beyond OOD itself including studying its properties as an uncertainty estimator in more general settings.

\acks{This work was performed under the auspices of the U.S. Department of Energy by Lawrence Livermore National Laboratory under Contract DE-AC52-07NA27344. Supported by the LDRD Program under projects 21-ERD-028, 22-ERD-006 and released under LLNL-JRNL-829478.}
\section*{Appendix}
We include experimental details, additional results, pseudo-code in the appendix. Code to reproduce our results is available at \href{https://github.com/LLNL/AMP}{github.com/LLNL/AMP}
\small{
\bibliography{refs}

\begin{thebibliography}{34}
\providecommand{\natexlab}[1]{#1}
\providecommand{\url}[1]{\texttt{#1}}
\expandafter\ifx\csname urlstyle\endcsname\relax
  \providecommand{\doi}[1]{doi: #1}\else
  \providecommand{\doi}{doi: \begingroup \urlstyle{rm}\Url}\fi

\bibitem[Arora et~al.(2019)Arora, Du, Hu, Li, and Wang]{arora2019fine}
Sanjeev Arora, Simon Du, Wei Hu, Zhiyuan Li, and Ruosong Wang.
\newblock Fine-grained analysis of optimization and generalization for
  overparameterized two-layer neural networks.
\newblock In \emph{International Conference on Machine Learning}, pages
  322--332. PMLR, 2019.

\bibitem[Bietti and Mairal(2019)]{bietti2019inductive}
Alberto Bietti and Julien Mairal.
\newblock On the inductive bias of neural tangent kernels.
\newblock \emph{Advances in Neural Information Processing Systems}, 32, 2019.

\bibitem[Blundell et~al.(2015)Blundell, Cornebise, Kavukcuoglu, and
  Wierstra]{blundell2015weight}
Charles Blundell, Julien Cornebise, Koray Kavukcuoglu, and Daan Wierstra.
\newblock Weight uncertainty in neural network.
\newblock In \emph{International Conference on Machine Learning}, pages
  1613--1622. PMLR, 2015.

\bibitem[Cimpoi et~al.(2014)Cimpoi, Maji, Kokkinos, Mohamed, and
  Vedaldi]{cimpoi2014describing}
Mircea Cimpoi, Subhransu Maji, Iasonas Kokkinos, Sammy Mohamed, and Andrea
  Vedaldi.
\newblock Describing textures in the wild.
\newblock In \emph{Proceedings of the IEEE Conference on Computer Vision and
  Pattern Recognition}, pages 3606--3613, 2014.

\bibitem[Gal and Ghahramani(2016)]{gal2016dropout}
Yarin Gal and Zoubin Ghahramani.
\newblock Dropout as a bayesian approximation: Representing model uncertainty
  in deep learning.
\newblock In \emph{international conference on machine learning}, pages
  1050--1059. PMLR, 2016.

\bibitem[Guo et~al.(2017)Guo, Pleiss, Sun, and Weinberger]{guo2017calibration}
Chuan Guo, Geoff Pleiss, Yu~Sun, and Kilian~Q Weinberger.
\newblock On calibration of modern neural networks.
\newblock In \emph{International Conference on Machine Learning}, pages
  1321--1330. PMLR, 2017.

\bibitem[He et~al.(2016)He, Zhang, Ren, and Sun]{he2016deep}
Kaiming He, Xiangyu Zhang, Shaoqing Ren, and Jian Sun.
\newblock Deep residual learning for image recognition.
\newblock In \emph{Proceedings of the IEEE conference on computer vision and
  pattern recognition}, pages 770--778, 2016.

\bibitem[Hendrycks and Dietterich(2019)]{hendrycks2018benchmarking}
Dan Hendrycks and Thomas Dietterich.
\newblock Benchmarking neural network robustness to common corruptions and
  perturbations.
\newblock In \emph{International Conference on Learning Representations}, 2019.
\newblock URL \url{https://openreview.net/forum?id=HJz6tiCqYm}.

\bibitem[Hendrycks et~al.(2019)Hendrycks, Mazeika, and
  Dietterich]{hendrycks18oe}
Dan Hendrycks, Mantas Mazeika, and Thomas Dietterich.
\newblock Deep anomaly detection with outlier exposure.
\newblock In \emph{ICLR}, 2019.

\bibitem[Jacot et~al.(2018)Jacot, Gabriel, and Hongler]{jacot2018neural}
Arthur Jacot, Franck Gabriel, and Cl{\'e}ment Hongler.
\newblock Neural tangent kernel: Convergence and generalization in neural
  networks.
\newblock \emph{Advances in neural information processing systems}, 31, 2018.

\bibitem[Jain et~al.(2021)Jain, Lahlou, Nekoei, Butoi, Bertin, Rector-Brooks,
  Korablyov, and Bengio]{jain2021deup}
Moksh Jain, Salem Lahlou, Hadi Nekoei, Victor Butoi, Paul Bertin, Jarrid
  Rector-Brooks, Maksym Korablyov, and Yoshua Bengio.
\newblock Deup: Direct epistemic uncertainty prediction.
\newblock \emph{arXiv preprint arXiv:2102.08501}, 2021.

\bibitem[Krishnan and Tickoo(2020)]{krishnan2020improving}
Ranganath Krishnan and Omesh Tickoo.
\newblock Improving model calibration with accuracy versus uncertainty
  optimization.
\newblock In H.~Larochelle, M.~Ranzato, R.~Hadsell, M.~F. Balcan, and H.~Lin,
  editors, \emph{Advances in Neural Information Processing Systems}, volume~33,
  pages 18237--18248. Curran Associates, Inc., 2020.
\newblock URL
  \url{https://proceedings.neurips.cc/paper/2020/file/d3d9446802a44259755d38e6d163e820-Paper.pdf}.

\bibitem[Krizhevsky et~al.(2009)Krizhevsky, Hinton,
  et~al.]{krizhevsky2009learning}
Alex Krizhevsky, Geoffrey Hinton, et~al.
\newblock Learning multiple layers of features from tiny images.
\newblock 2009.

\bibitem[Lakshminarayanan et~al.(2017)Lakshminarayanan, Pritzel, and
  Blundell]{lakshminarayanan2016simple}
Balaji Lakshminarayanan, Alexander Pritzel, and Charles Blundell.
\newblock Simple and scalable predictive uncertainty estimation using deep
  ensembles.
\newblock \emph{Advances in neural information processing systems}, 30, 2017.

\bibitem[Lee et~al.(2019)Lee, Xiao, Schoenholz, Bahri, Novak, Sohl-Dickstein,
  and Pennington]{lee2019wide}
Jaehoon Lee, Lechao Xiao, Samuel Schoenholz, Yasaman Bahri, Roman Novak, Jascha
  Sohl-Dickstein, and Jeffrey Pennington.
\newblock Wide neural networks of any depth evolve as linear models under
  gradient descent.
\newblock \emph{Advances in neural information processing systems}, 32, 2019.

\bibitem[Lee et~al.(2018{\natexlab{a}})Lee, Lee, Lee, and Shin]{lee2018simple}
Kimin Lee, Kibok Lee, Honglak Lee, and Jinwoo Shin.
\newblock A simple unified framework for detecting out-of-distribution samples
  and adversarial attacks.
\newblock \emph{Advances in neural information processing systems}, 31,
  2018{\natexlab{a}}.

\bibitem[Lee et~al.(2018{\natexlab{b}})Lee, Lee, Lee, and Shin]{mahalanobis}
Kimin Lee, Kibok Lee, Honglak Lee, and Jinwoo Shin.
\newblock A simple unified framework for detecting out-of-distribution samples
  and adversarial attacks.
\newblock In \emph{NeurIPS}, 2018{\natexlab{b}}.

\bibitem[Liang et~al.(2017)Liang, Li, and Srikant]{odin}
Shiyu Liang, Yixuan Li, and Rayadurgam Srikant.
\newblock Enhancing the reliability of out-of-distribution image detection in
  neural networks.
\newblock In \emph{ICLR}, 2017.

\bibitem[Liu et~al.(2020)Liu, Wang, Owens, and Li]{energyood}
Weitang Liu, Xiaoyun Wang, John Owens, and Yixuan Li.
\newblock Energy-based out-of-distribution detection.
\newblock In \emph{NeurIPS}, 2020.

\bibitem[Neal(2012)]{neal2012bayesian}
Radford~M Neal.
\newblock \emph{Bayesian learning for neural networks}, volume 118.
\newblock Springer Science \& Business Media, 2012.

\bibitem[Netzer et~al.(2011)Netzer, Wang, Coates, Bissacco, Wu, and
  Ng]{netzer2011reading}
Yuval Netzer, Tao Wang, Adam Coates, Alessandro Bissacco, Bo~Wu, and Andrew~Y
  Ng.
\newblock Reading digits in natural images with unsupervised feature learning.
\newblock 2011.

\bibitem[Ovadia et~al.(2019)Ovadia, Fertig, Ren, Nado, Sculley, Nowozin,
  Dillon, Lakshminarayanan, and Snoek]{ovadia2019}
Yaniv Ovadia, Emily Fertig, Jie Ren, Zachary Nado, David Sculley, Sebastian
  Nowozin, Joshua Dillon, Balaji Lakshminarayanan, and Jasper Snoek.
\newblock Can you trust your model's uncertainty? evaluating predictive
  uncertainty under dataset shift.
\newblock \emph{Advances in neural information processing systems}, 32, 2019.

\bibitem[Parmar et~al.(2021)Parmar, Zhang, and Zhu]{parmar2021buggy}
Gaurav Parmar, Richard Zhang, and Jun-Yan Zhu.
\newblock On buggy resizing libraries and surprising subtleties in fid
  calculation.
\newblock \emph{arXiv preprint arXiv:2104.11222}, 2021.

\bibitem[Russakovsky et~al.(2015)Russakovsky, Deng, Su, Krause, Satheesh, Ma,
  Huang, Karpathy, Khosla, Bernstein, et~al.]{russakovsky2015imagenet}
Olga Russakovsky, Jia Deng, Hao Su, Jonathan Krause, Sanjeev Satheesh, Sean Ma,
  Zhiheng Huang, Andrej Karpathy, Aditya Khosla, Michael Bernstein, et~al.
\newblock Imagenet large scale visual recognition challenge.
\newblock \emph{International journal of computer vision}, 115\penalty0
  (3):\penalty0 211--252, 2015.

\bibitem[Sastry and Oore(2020)]{sastry2020detecting}
Chandramouli~Shama Sastry and Sageev Oore.
\newblock Detecting out-of-distribution examples with gram matrices.
\newblock In \emph{International Conference on Machine Learning}, pages
  8491--8501. PMLR, 2020.

\bibitem[Talebi and Milanfar(2021)]{talebi2021learning}
Hossein Talebi and Peyman Milanfar.
\newblock Learning to resize images for computer vision tasks.
\newblock In \emph{Proceedings of the IEEE/CVF International Conference on
  Computer Vision}, pages 497--506, 2021.

\bibitem[Tancik et~al.(2020)Tancik, Srinivasan, Mildenhall, Fridovich-Keil,
  Raghavan, Singhal, Ramamoorthi, Barron, and Ng]{tancik2020fourier}
Matthew Tancik, Pratul Srinivasan, Ben Mildenhall, Sara Fridovich-Keil, Nithin
  Raghavan, Utkarsh Singhal, Ravi Ramamoorthi, Jonathan Barron, and Ren Ng.
\newblock Fourier features let networks learn high frequency functions in low
  dimensional domains.
\newblock \emph{Advances in Neural Information Processing Systems},
  33:\penalty0 7537--7547, 2020.

\bibitem[Van~Amersfoort et~al.(2020)Van~Amersfoort, Smith, Teh, and
  Gal]{van2020uncertainty}
Joost Van~Amersfoort, Lewis Smith, Yee~Whye Teh, and Yarin Gal.
\newblock Uncertainty estimation using a single deep deterministic neural
  network.
\newblock In \emph{International Conference on Machine Learning}, pages
  9690--9700. PMLR, 2020.

\bibitem[Xu et~al.(2015)Xu, Ehinger, Zhang, Finkelstein, Kulkarni, and
  Xiao]{xu2015turkergaze}
Pingmei Xu, Krista~A Ehinger, Yinda Zhang, Adam Finkelstein, Sanjeev~R
  Kulkarni, and Jianxiong Xiao.
\newblock Turkergaze: Crowdsourcing saliency with webcam based eye tracking.
\newblock \emph{arXiv preprint arXiv:1504.06755}, 2015.

\bibitem[Yang et~al.(2021)Yang, Wang, Feng, Yan, Zheng, Zhang, and Liu]{scood}
Jingkang Yang, Haoqi Wang, Litong Feng, Xiaopeng Yan, Huabin Zheng, Wayne
  Zhang, and Ziwei Liu.
\newblock Semantically coherent out-of-distribution detection.
\newblock In \emph{Proceedings of the IEEE/CVF International Conference on
  Computer Vision}, pages 8301--8309, 2021.

\bibitem[Yu et~al.(2015)Yu, Seff, Zhang, Song, Funkhouser, and
  Xiao]{yu2015lsun}
Fisher Yu, Ari Seff, Yinda Zhang, Shuran Song, Thomas Funkhouser, and Jianxiong
  Xiao.
\newblock Lsun: Construction of a large-scale image dataset using deep learning
  with humans in the loop.
\newblock \emph{arXiv preprint arXiv:1506.03365}, 2015.

\bibitem[Yu and Aizawa(2019)]{mcd}
Qing Yu and Kiyoharu Aizawa.
\newblock Unsupervised out-of-distribution detection by maximum classifier
  discrepancy.
\newblock In \emph{ICCV}, 2019.

\bibitem[Zagoruyko and Komodakis(2016)]{zagoruyko2016wide}
Sergey Zagoruyko and Nikos Komodakis.
\newblock Wide residual networks.
\newblock \emph{arXiv preprint arXiv:1605.07146}, 2016.

\bibitem[Zhou et~al.(2017)Zhou, Lapedriza, Khosla, Oliva, and
  Torralba]{zhou2017places}
Bolei Zhou, Agata Lapedriza, Aditya Khosla, Aude Oliva, and Antonio Torralba.
\newblock Places: A 10 million image database for scene recognition.
\newblock \emph{IEEE transactions on pattern analysis and machine
  intelligence}, 40\penalty0 (6):\penalty0 1452--1464, 2017.

\end{thebibliography}
}
\newpage
\appendix
\section{Details of Benchmark Datasets}
 We use commonly used benchmarks to evaluate AMP, these include the following datasets -- iSUN \citep{xu2015turkergaze}, LSUN (R), LSUN (C) \citep{yu2015lsun}, Places365 \citep{zhou2017places}, Texture \citep{cimpoi2014describing}, and SVHN \citep{netzer2011reading}
\paragraph{Consistency training details} The transformation $\mathcal{T}$ was applied to the anchors using a pre-specified schedule, every $5^{\text{th}}$ batch for CIFAR-10/100 and every $10^{\text{th}}$ batch for ImageNet, while the clean anchors were used directly in the other batches. However, from our experiments, we found that the choice of this schedule is not sensitive and the detection performance was similar even with other schedules. During the inference step, we did not utilize any transformation $\mathcal{T}$, and fixed the number of anchors $K=5$ while making predictions for a test image. We performed an ablation on the number of anchors (reported at the end of the section), and observed that even a small number of random anchors was sufficient to obtain good detection performance, thus making our approach efficient in practice. 

\paragraph{}During training we always use $K=1$ anchor, which is typically chosen by randomly shuffling the current batch so that every input sample is assigned a random anchor from that batch. During training we use \texttt{RandomCrop, RandomHorizontalFlip} augmentations in Pytorch. For the test set and the OOD set, we normalize data to the same mean and standard deviation as the training set without any additional transformations.

\section{Hyperparameter settings}
\noindent \textbf{CIFAR-10/100:} We use standard training protocol for both CIFAR-10/100 datasets using all our networks -- WideResNet, ResNet-18, ResNet-34 \citep{he2016deep}.  We use an SGD optimizer with an initial learning rate of 0.1, momentum of 0.9, and weight decay of $5e-4$. This learning rate is scaled down by a $\gamma=0.2$ using a schedule of [60, 120, 160] epochs out of the total 200 epochs for training.  We use a batch size of 128 in all our training experiments for CIFAR datasets.
\noindent \textbf{ImageNet:} We also follow standard training protocol for ResNet-50 on ImageNet as well. We use an SGD optimizer with a learning rate of 0.1, weight decay of $1e-4$, momentum of 0.9. We decay the learning rate by 0.1 every 30 epochs, and train for a total of 120 epochs. We use a batch size of 128 to train the model.
\begin{table}[!htb]
\small
 \centering

		\begin{tabular}{l|l}
			\toprule
        Method & {AUROC $\uparrow$}  \\ \midrule
		 ResNet-18 \citep{he2016deep} & 91.77 $\pm$ 1.85\\
		 DUQ \citep{van2020uncertainty} & 92.70 $\pm$ 1.30\\
		 Deep Ens \citep{lakshminarayanan2016simple} & 94.70\\
	 	\midrule
	 	\name  & \textbf{97.41 $\pm$ 0.72} \\
	    \bottomrule
		\end{tabular}
\caption{\small{OOD Detection with uncertainties on CIFAR-SVHN with ResNet-18.}}
\label{tab:cifar-svhn}
\end{table}

\section{Modification to anchor a model}
We demonstrate in \ref{alg:anchoring}, the simple modification to be able to train with anchoring
\begin{algorithm}[tb]
\caption{PyTorch-style pseudo-code for anchoring.}
\label{alg:anchoring}
\begin{center}
\begin{python}
def create_anchored_model(model):
    model.conv1 = nn.Conv2d(in_channels=6, 64)
    return model
    
Tx = transforms.Compose([
    transforms.RandomResizedCrop(size=224),
    transforms.RandomHorizontalFlip(),
    transforms.RandomApply([color_jitter,blurr], p=0.8),
    ])    
## load model and change the first conv layer

model_basic = ResNet50(pre_trained=False,n_class=1000)
model = create_anchored_model(model_basic)

## load datasets, setup optimizer, define criterion etc.
for images, targets in train_loder:
    batch_order = np.arange(images.shape[0])
    np.random.shuffle(batch_order)
    anchors = images[batch_order,:,:,:]
    diff = images-anchors
    if i % 10 ==0:
        tx_anchors = Tx(anchors)
    else:
        tx_anchors = anchors

    batch = torch.cat([tx_anchors,diff],axis=1)
    output = model(batch)
    loss = criterion(output, target)
    optimizer.zero_grad()
    loss.backward()
    optimizer.step()
\end{python}
\end{center}
\end{algorithm}

\section{Additional Results}
We report detailed results for individual datasets on various benchmarks used in the paper here. \Cref{tab:detailed_cifar100_res18} and \Cref{tab:detailed_cifar10_res18} report 4 performance metrics for the SCOOD benchmark \citep{scood}, where we use the re-sampled OOD set following the SCOOD protocol. We observe competitive performance on CIFAR-10 and state-of-the-art on CIFAR-100 with AMP. Next, \Cref{tab:ood_results-res34}, we report detailed performance numbers on the second OOD benchmark used in the paper. We note that our method consistently performs either the best or second best as compared to GM \citep{sastry2020detecting}, while being better on average across the various datasets. In particular, we see that on challenging datasets like near-OOD AMP is significantly better than all competing baselines.  Finally, in \Cref{tab:cifar-svhn} we show uncertainty based OOD on a CIFAR-10 vs SVHN benchmark, compared to other uncertainty based approaches. We see once again that AMP is significantly better than sophisticated methods including Deep Ensembles that requires multiple models to be trained.

\begin{table*}[btp!]
\centering
\footnotesize
\begin{tabular}{c|c|ccc}
\toprule
\multirow{1}{*}{Method} & \multirow{1}{*}{Dataset} 
& \multirow{1}{*}{FPR95~$\downarrow$} 
& \multirow{1}{*}{AUROC~$\uparrow$} 
& \multirow{1}{*}{AUPR(In/Out)~$\uparrow$}\\


\midrule
\multirow{7}{*}{\begin{tabular}[c]{@{}c@{}}ODIN\end{tabular}}

&Texture     & 42.52          & 84.06          & 86.01          ~/~ 80.73       \\
&SVHN     & 52.27          & 83.26          & 63.76          ~/~ 92.60          \\
& CIFAR-100         & 56.34          & 78.40          & 73.21          ~/~ 80.99    \\
& Tiny-ImageNet & 59.09          & 79.69          & 79.34          ~/~ 77.52    \\
& LSUN &  47.85          & 84.56          & 81.56          ~/~ 85.58          \\
& Places365 &  53.94          & 82.01          & 54.92          ~/~ 93.30          \\
\cmidrule{2-5}
& \textbf{Mean} & \textbf{52.00} & \textbf{82.00} & \textbf{73.13} ~/~ \textbf{85.12} \\

\midrule
\multirow{7}{*}{\begin{tabular}[c]{@{}c@{}}EBO\end{tabular}}

&Texture     & 52.11          & 80.70          & 83.34          ~/~ 75.20       \\
&SVHN     & 30.56          & 92.08          & 80.95          ~/~ 96.28          \\
& CIFAR-100         & 56.98          & 79.65          & 75.09          ~/~ 81.23      \\
& Tiny-ImageNet & 57.81          & 81.65          & 81.80          ~/~ 78.75          \\
& LSUN &  50.56          & 85.04          & 82.80          ~/~ 85.29          \\
& Places365 &  52.16          & 83.86          & 58.96          ~/~ 93.90          \\
\cmidrule{2-5}
& \textbf{Mean} & \textbf{50.03} & \textbf{83.83} & \textbf{77.15} ~/~ \textbf{85.11} \\

\midrule
\multirow{7}{*}{\begin{tabular}[c]{@{}c@{}}MCD\end{tabular}}

&Texture     & 83.92          & 81.59          & 90.20           ~/~ 63.27      \\
&SVHN     & 60.27          & 89.78          & 85.33          ~/~ 94.25          \\
& CIFAR-100         & 74.00             & 82.78          & 83.97          ~/~ 79.16         \\
& Tiny-ImageNet & 78.89          & 80.98          & 85.63          ~/~ 72.48          \\
& LSUN &  68.96          & 84.71          & 85.74          ~/~ 81.50          \\
& Places365 &  72.08          & 83.51          & 69.44          ~/~ 92.52         \\
\cmidrule{2-5}
& \textbf{Mean} & \textbf{73.02} & \textbf{83.89} & \textbf{83.39} ~/~ \textbf{80.53} \\

\midrule
\multirow{7}{*}{\begin{tabular}[c]{@{}c@{}}OE\end{tabular}}

&Texture     & 51.17          & 89.56          & 93.79          ~/~ 81.88       \\
&SVHN     & 20.88          & 96.43          & 93.62          ~/~ 98.32          \\
& CIFAR-100         & 58.54          & 86.22          & 86.17          ~/~ 84.88         \\
& Tiny-ImageNet & 58.98          & 87.65          & 90.9           ~/~ 82.16         \\
& LSUN &  57.97          & 86.75          & 87.69          ~/~ 85.07         \\
& Places365 &  55.64          & 87.00             & 73.11          ~/~ 94.67          \\
\cmidrule{2-5}
& \textbf{Mean} & \textbf{50.53} & \textbf{88.93} & \textbf{87.55} ~/~ \textbf{87.83} \\

\midrule
\multirow{7}{*}{\begin{tabular}[c]{@{}c@{}}UDG\end{tabular}}

&Texture     & 20.43          & 96.44          & 98.12          ~/~ 92.91      \\
&SVHN     & 13.26          & 97.49          & 95.66          ~/~ 98.69         \\
& CIFAR-100         & 47.20           & 90.98          & 91.74          ~/~ 89.36    \\
& Tiny-ImageNet & 50.18          & 91.91          & 94.43          ~/~ 86.99         \\
& LSUN &  42.05          & 93.21          & 94.53          ~/~ 91.03        \\
& Places365 &  44.22          & 92.64          & 87.17          ~/~ 96.66   \\
\cmidrule{2-5}
& \textbf{Mean} & \textbf{36.22} & \textbf{93.78} & \textbf{93.61} ~/~ \textbf{92.61} \\
\midrule
\multirow{7}{*}{\begin{tabular}[c]{@{}c@{}}AMP (ours)\end{tabular}}

&Texture     & 52.43          & 88.74          & 91.91          ~/~ 80.48      \\
&SVHN     & 12.53          & 97.60          & 95.58          ~/~ 98.83         \\
& CIFAR-100         & 48.10           & 89.61          & 88.99          ~/~ 88.47    \\
& Tiny-ImageNet & 50.40          & 90.26          & 92.01          ~/~ 85.74         \\
& LSUN &  23.01          & 95.17          & 94.94          ~/~ 94.78        \\
& Places365 &  34.45          & 93.25          & 83.95          ~/~ 97.19   \\
\cmidrule{2-5}
& \textbf{Mean} & \textbf{36.82} & \textbf{92.40} & \textbf{91.23} ~/~ \textbf{90.91} \\

\bottomrule
\end{tabular}
\caption{\textbf{Detailed results on SCOOD benchmark \citep{scood} using CIFAR-10/ResNet-18.} AMP performs very close to methods that use outlier exposure, while outperforming all the baselines that do not. We use results for baselines as reported in \citep{scood}}
\label{tab:detailed_cifar10_res18}
\end{table*}

\begin{table*}[btp!]
\setlength{\tabcolsep}{6pt}
\footnotesize
\centering
\begin{tabular}{c|c|ccc}
\toprule
\multirow{1}{*}{Method} & \multirow{1}{*}{Dataset} 
& \multirow{1}{*}{FPR95~$\downarrow$} 
& \multirow{1}{*}{AUROC~$\uparrow$} 
& \multirow{1}{*}{AUPR(In/Out)~$\uparrow$}\\
\midrule

\multirow{7}{*}{\begin{tabular}[c]{@{}c@{}}ODIN\end{tabular}}

&Texture     & 79.47          & 77.92          & 86.69          ~/~ 62.97           \\
&SVHN     & 90.33          & 75.59          & 65.25          ~/~ 84.49         \\
& CIFAR-10    & 81.82          & 77.90          & 79.93          ~/~ 73.39     \\
& Tiny-ImageNet & 82.74          & 77.58          & 86.26          ~/~ 61.38    \\
& LSUN &  80.57          & 78.22          & 86.34          ~/~ 63.44         \\
& Places365 &  76.42          & 80.66          & 66.77          ~/~ 89.66    \\
\cmidrule{2-5}
& \textbf{Mean} & \textbf{81.89} & \textbf{77.98} & \textbf{78.54} ~/~ \textbf{72.56} \\
\midrule
\multirow{7}{*}{\begin{tabular}[c]{@{}c@{}}EBO\end{tabular}}

& Texture     & 84.29          & 76.32          & 85.87          ~/~ 59.12          \\
&SVHN     & 78.23          & 83.57          & 75.61          ~/~ 90.24          \\
& CIFAR-10    & 81.25          & 78.95          & 80.01          ~/~ 74.44      \\
& Tiny-ImageNet & 83.32          & 78.34          & 87.08          ~/~ 62.13    \\
& LSUN &  84.51          & 77.66          & 86.42          ~/~ 61.40          \\
& Places365 &  78.37          & 80.99          & 68.22          ~/~ 89.60     \\
\cmidrule{2-5}
& \textbf{Mean} & \textbf{81.66} & \textbf{79.31} & \textbf{80.54} ~/~ \textbf{72.82} \\
\midrule  
\multirow{7}{*}{\begin{tabular}[c]{@{}c@{}}MCD\end{tabular}}
&Texture     & 83.97          & 73.46          & 83.11          ~/~ 56.79        \\
&SVHN     & 85.82          & 76.61          & 65.50          ~/~ 85.52          \\
& CIFAR-10    & 87.74          & 73.15          & 76.51          ~/~ 67.24      \\
& Tiny-ImageNet  & 84.46          & 75.32          & 85.11          ~/~ 59.49   \\
& LSUN & 86.08          & 74.05          & 84.21          ~/~ 58.62         \\
& Places365 & 82.74          & 76.30          & 61.15          ~/~ 87.19    \\
\cmidrule{2-5}
& \textbf{Mean} & \textbf{85.14} & \textbf{74.82} & \textbf{75.93} ~/~ \textbf{69.14} \\
\midrule
\multirow{7}{*}{\begin{tabular}[c]{@{}c@{}}OE\end{tabular}}
&Texture     & 86.56          & 73.89          & 84.48          ~/~ 54.84         \\
&SVHN     & 68.87          & 84.23          & 75.11          ~/~ 91.41        \\
& CIFAR-10    & 79.72          & 78.92          & 81.95          ~/~ 74.28    \\
& Tiny-ImageNet & 83.41          & 76.99          & 86.36          ~/~ 60.56   \\
& LSUN &  83.53          & 77.10          & 86.28          ~/~ 60.97          \\
& Places365 &  78.24          & 79.62          & 67.13          ~/~ 88.89     \\
\cmidrule{2-5}
& \textbf{Mean} & \textbf{80.06} & \textbf{78.46} & \textbf{80.22} ~/~ \textbf{71.83} \\
\midrule
\multirow{7}{*}{\begin{tabular}[c]{@{}c@{}}UDG\end{tabular}}
&Texture     & 75.04          & 79.53          & 87.63          ~/~ 65.49         \\
&SVHN     & 60.00          & 88.25          & 81.46          ~/~ 93.63        \\
& CIFAR-10    & 83.35          & 76.18          & 78.92          ~/~ 71.15    \\
& Tiny-ImageNet & 81.73          & 77.18          & 86.00          ~/~ 61.67  \\
& LSUN &  78.70          & 76.79          & 84.74          ~/~ 63.05         \\
& Places365 &  73.86          & 79.87          & 65.36          ~/~ 89.60    \\
\cmidrule{2-5}
& \textbf{Mean} & \textbf{75.45} & \textbf{79.63} & \textbf{80.69} ~/~ \textbf{74.10}\\
\midrule
\multirow{7}{*}{\begin{tabular}[c]{@{}c@{}}AMP  (ours)\end{tabular}}
&Texture     & 68.39          & 83.76          & 90.69          ~/~ 72.16         \\
&SVHN     & 34.12          & 94.21          & 90.11          ~/~ 97.24        \\
& CIFAR-10    & 80.47          & 78.74          & 81.36          ~/~ 74.07    \\
& Tiny-ImageNet & 80.70          & 78.34          & 86.95          ~/~ 63.03  \\
& LSUN &  83.60          & 76.64          & 85.80          ~/~ 60.63         \\
& Places365 &  74.77          & 81.67          & 69.97          ~/~ 90.09    \\
\cmidrule{2-5}
& \textbf{Mean} & \textbf{70.34} & \textbf{82.22} & \textbf{84.14} ~/~ \textbf{76.20}\\
\bottomrule
\end{tabular}
\caption{\textbf{Detailed results on SCOOD benchmark \citep{scood} using CIFAR-100/ResNet-18.} AMP consistently outperforms all methods including those that use outlier exposure. We use results for baselines as reported in \citep{scood}}
\label{tab:detailed_cifar100_res18}
\end{table*}

\begin{table*}[htbp]
\footnotesize
\resizebox{\linewidth}{!}{
\centering
\begin{tabular}{cllll}
\hline
\multirow{2}{*}{\begin{tabular}[c]{@{}c@{}}In-dist\\(model)\end{tabular}}      & \multicolumn{1}{c}{\multirow{2}{*}{OOD}} & \multicolumn{1}{c}{TNR at TPR 95\% $\uparrow$}         & \multicolumn{1}{c}{AUROC $\uparrow$}                   & \multicolumn{1}{c}{Detection Acc. $\uparrow$}           \\
\cline{3-5}
                                                                              & \multicolumn{1}{c}{}                     & \multicolumn{3}{c}{MSP / ODIN / Gram Matrices / Ours}                                                                                 \\
\hline
\multirow{7}{*}{\begin{tabular}[c]{@{}c@{}}CIFAR-10\\(ResNet-34)\end{tabular}}  & iSUN             & 44.6 / 73.2 / \textbf{97.3} / 91.8     & 91.0 / 94.0 / \textbf{99.1} / 98.2    & 85.0 / 86.5 / \textbf{96.2} / 93.8           \\
                                                                                & LSUN (R)         & 49.8 / 82.1 / \textbf{98.2} / 92.4     & 91.0 / 94.1 / \textbf{99.2} / 98.7    & 85.3 / 86.7 / \textbf{96.7} / 94.9          \\
                                                                                & LSUN (C)        & 48.6 / 62.0 / 91.7 / \textbf{98.5}     & 91.9 / 91.2 / 98.3 / \textbf{99.5}    & 86.3 / 82.4 / 94.1 / \textbf{97.0}           \\
                                                                                & TinyImgNet (R)  & 41.0 / 67.9 / \textbf{95.9} / 88.8     & 91.0 / 94.0 / \textbf{98.9} / 97.0    & 85.1 / 86.5 / \textbf{95.6} / 92.1           \\
                                                                                & TinyImgNet (C)  & 46.4 / 68.7 / 77.6 / \textbf{94.5}     & 91.4 / 93.1 / 96.2 / \textbf{98.7}    & 85.4 / 85.2 / 90.8 / \textbf{94.9}          \\
                                                                                & SVHN            & 50.5 / 70.3 / \textbf{95.3} / 91.2     & 89.9 / 96.7 / \textbf{99.0} / 98.1    & 85.1 / 91.1 / \textbf{95.2} / 93.7           \\
                                                                                & CIFAR-100       & 33.3 / 42.0 / 40.2 / \textbf{56.5}     & 86.4 / 85.8 / 83.6 / \textbf{90.2}    & 80.4 / 78.6 / 76.4 / \textbf{83.5} \\
\hline
\multirow{7}{*}{\begin{tabular}[c]{@{}c@{}}CIFAR-100\\(ResNet-34)\end{tabular}}& iSUN        & 16.9 / 45.2 / \textbf{66.2} / 48.7         & 75.8 / 85.5 /  \textbf{94.6} / 90.2          & 70.1 / 78.5 / \textbf{88.3} / 82.6           \\
                                                                              & LSUN (R)           & 18.8 / 23.2 / \textbf{61.4} / 54.2         & 75.8 / 85.6 / \textbf{ 94.4} / 91.7          & 69.9 / 78.3 / \textbf{88.6} / 84.3           \\
                                                                              & LSUN (C)           & 18.7 / 44.1 / 43.7 / \textbf{67.8}         & 75.5 / 82.7 / 89.7 / \textbf{94.0}           & 69.2 / 75.9 / 82.4 / \textbf{86.4}           \\
                                                                              & TinyImgNet (R)     & 20.4 / 36.1 / \textbf{66.8} / 45.9         & 77.2 / 87.6 / \textbf{94.7} / 89.2           & 70.8 / 80.1 / \textbf{88.6} / 81.3           \\
                                                                              & TinyImgNet (C)     & 24.3 / 44.3 / 41.4 / \textbf{61.5}         & 79.7 / 85.4 /  89.7 / \textbf{92.9}          & 72.5 / 78.3 / 82.8 / \textbf{85.4}           \\
                                                                              & SVHN               & 20.3 / \textbf{62.7} / 54.5 / 56.5         & 79.5 / \textbf{93.9} / 92.1 / 91.9           & 73.2 / \textbf{88.0} / 84.9 / 83.7          \\
                                                                              & CIFAR-10           & \textbf{19.1} / 18.7 / 16.9 / 17.5         & 77.1 / 77.2 /  74.5  / \textbf{79.9}         & 71.0 / 71.2 /  68.9 / \textbf{73.8}          \\
\hline
\end{tabular}
}
\caption{Detailed results on the OOD detection benchmark with ResNet-34. Note, different from the main paper we report TNR here (instead of FPR95) which is 100-FPR95, as this was used in \citep{sastry2020detecting}. We observe that AMP performs comparably to Gram Matrices, while being better on average. Our method has significant advantages on more challenging datasets like near OOD.}
\label{tab:ood_results-res34}
\end{table*}

\end{document}